\documentclass{article}

\usepackage[preprint]{neurips_2025}

\usepackage{graphicx}
\usepackage{amsmath}
\usepackage{amssymb}
\usepackage{booktabs}
\usepackage{algorithm}
\usepackage{algpseudocode}
\usepackage{xspace}

\usepackage{amsmath,amsfonts,bm}

\def\eqref#1{equation~\ref{#1}}

\def\1{\bm{1}}

\def\mA{{\bm{A}}}

\def\mE{{\bm{E}}}

\def\mI{{\bm{I}}}

\def\mM{{\bm{M}}}

\def\mP{{\bm{P}}}
\def\mQ{{\bm{Q}}}

\def\mS{{\bm{S}}}

\DeclareMathAlphabet{\mathsfit}{\encodingdefault}{\sfdefault}{m}{sl}
\SetMathAlphabet{\mathsfit}{bold}{\encodingdefault}{\sfdefault}{bx}{n}

\def\sR{{\mathbb{R}}}

\newcommand{\Ls}{\mathcal{L}}

\usepackage{amsthm}
\usepackage{tablefootnote}
\usepackage{multirow}
\usepackage{color, colortbl}
\usepackage{pifont}

\makeatletter
\DeclareRobustCommand\onedot{\futurelet\@let@token\@onedot}
\def\@onedot{\ifx\@let@token.\else.\null\fi\xspace}

\def\ie{\emph{i.e}\onedot}

\def\etal{\emph{et al}\onedot}
\makeatother

\definecolor{mygray}{RGB}{238, 238, 238}
\definecolor{myblue}{RGB}{218, 232, 252}
\definecolor{myred}{RGB}{255, 204, 204}
\definecolor{mygreen}{RGB}{205, 235, 139}
\definecolor{myorange}{RGB}{250, 215, 172}
\definecolor{myyellow}{RGB}{255, 255, 136}

\usepackage[utf8]{inputenc} 
\usepackage[T1]{fontenc}    
\usepackage{hyperref}       
\usepackage{url}            
\usepackage{booktabs}       
\usepackage{amsfonts}       
\usepackage{nicefrac}       
\usepackage{microtype}      
\usepackage{xcolor}         
\usepackage{amsmath}
\usepackage{multirow}
\usepackage{graphicx}
\usepackage{cleveref}

\title{\texttt{iPad}: Iterative Proposal-centric End-to-End Autonomous Driving}

\author{%
  Ke Guo\\
  Nanyang Technological University\\
  \texttt{ke.guo@ntu.edu.sg} \\
  \And
  Haochen Liu \\
  Nanyang Technological University \\
  \texttt{haochen002@e.ntu.edu.sg} \\
  \AND
  Xiaojun Wu \\
  Desay SV Automotive \\
  \texttt{Xiaojun.Wu@desaysv.com} \\
  \And
  Jia Pan \\
  The University of Hong Kong \\
  \texttt{jpan@cs.hku.hk} \\
  \And
   Chen Lv \\
  Nanyang Technological University \\
  \texttt{lyuchen@ntu.edu.sg} \\
}

\begin{document}

\maketitle

\begin{abstract}
End-to-end (E2E) autonomous driving systems offer a promising alternative to traditional modular pipelines by reducing information loss and error accumulation, with significant potential to enhance both mobility and safety. However, most existing E2E approaches directly generate plans based on dense bird’s-eye view (BEV) grid features, leading to inefficiency and limited planning awareness. To address these limitations, we propose iterative Proposal-centric autonomous driving (\texttt{iPad}), a novel framework that places proposals—a set of candidate future plans—at the center of feature extraction and auxiliary tasks. Central to \texttt{iPad} is ProFormer, a BEV encoder that iteratively refines proposals and their associated features through proposal-anchored attention, effectively fusing multi-view image data. Additionally, we introduce two lightweight, proposal-centric auxiliary tasks—mapping and prediction—that improve planning quality with minimal computational overhead. Extensive experiments on the NAVSIM and CARLA Bench2Drive benchmarks demonstrate that \texttt{iPad} achieves state-of-the-art performance while being significantly more efficient than prior leading methods. Code is available at \url{https://github.com/Kguo-cs/iPad}.
\end{abstract}

\section{Introduction}
\label{sec:intro}
Autonomous vehicles have garnered significant research interest due to their potential to revolutionize transportation and enhance traffic safety~\cite{yurtsever2020survey}. Traditional autonomous driving systems are typically composed of modular components—localization, perception, tracking, prediction, planning, and control—to ensure interpretability. However, the decoupled learning and design across these modules often lead to information loss and error accumulation. Recently, end-to-end (E2E) driving paradigms have emerged as a promising alternative~\cite{chen2024end}, leveraging holistic, fully differentiable models that map raw sensor data directly to planning outputs.

\begin{figure}[t]
\begin{center}
\includegraphics[width=\linewidth]{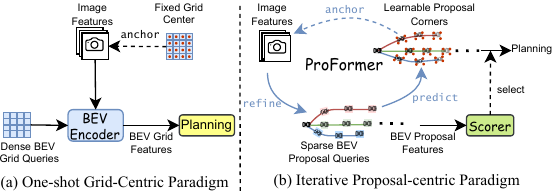}
\end{center}
\vspace{-0.3cm}
\caption{\textbf{Comparison of end-to-end paradigms.} (a) Dense one-shot, grid-centric methods generate BEV features for every cell and directly output the final plan based on the extracted dense BEV grid features.  (b) \texttt{iPad} iteratively refines sparse BEV proposals and their queries, concentrating feature extraction on the regions most relevant to planning by using the proposal corner points as anchors.}
\vspace{-1.0cm}
\label{fig:introduction}
\end{figure}

Early E2E approaches such as ALVINN~\cite{pomerleau1988alvinn} and PilotNet~\cite{bojarski2016end} aimed to learn a direct mapping from high-dimensional inputs to trajectories or control commands. However, these straightforward models were difficult to optimize and lacked interpretability. To address these shortcomings, more recent work~\cite{hu2023planning,jiang2023vad,chen2024vadv2,chen2025ppad,liu2025hybrid} introduces intermediate BEV grid features using a BEV encoder~\cite{philion2020lift,li2022bevformer} to fuse multi-view image features, as shown in~\cref{fig:introduction}, which are then used to directly generate final driving plans. While BEV-based pipelines improve interpretability, their dense grids incur substantial computational cost~\cite{jia2025drivetransformer} and often capture spurious correlations with irrelevant scene elements—leading to degraded planning performance and causal confusion~\cite{codevilla2019exploring,li2024ego}.

To overcome these limitations, we propose \textbf{iterative Proposal-centric autonomous driving (\texttt{iPad})}, a unified E2E framework that places proposals at the heart of the model. In \texttt{iPad}, each proposal is a candidate future trajectory, and feature extraction, mapping, prediction, and scoring are all centered around these sparse BEV proposals. Unlike prior work that treats planning as a final-stage task built on fixed intermediate features, \texttt{iPad} makes planning the central organizing principle of the entire architecture. Specifically, it formulates planning as an iterative process of proposal refinement. We begin by initializing BEV proposal queries based on the ego vehicle’s current state. We then introduce \textbf{ProFormer}, a proposal-centric BEV encoder that predicts proposals from these queries. Using the corner points of all proposals as anchor points, multi-view image features are aggregated around them to refine the proposal queries. This predict–anchor–refine cycle repeats iteratively, producing increasingly accurate proposals and BEV proposal features. Finally, a lightweight scoring module evaluates the refined proposals and selects the best trajectory for execution.

\texttt{iPad} excels in both efficiency and effectiveness. In terms of efficiency, it scales linearly with the number of proposals, in contrast to the quadratic complexity of dense BEV grid methods. By employing planning-aware image feature extraction, \texttt{iPad} directly captures task-relevant information—avoiding the information bottlenecks inherent to dense grid representations. Furthermore, by modeling multi-modal expert planning distributions with a diverse set of learnable proposals, \texttt{iPad} can mitigate the modal collapse common in widely used deterministic planners such as Transfuser~\cite{chitta2022transfuser}, ST-P3~\cite{hu2022st}, and UniAD~\cite{hu2023planning}.

In addition, most existing E2E methods incorporate auxiliary tasks—such as object detection~\cite{sun2024sparsedrive}, occupancy prediction~\cite{hu2023planning}, or motion forecasting~\cite{jiang2023vad}—to enhance intermediate representations learning. However, these often need dense, computationally expensive features and are poorly aligned with the ultimate planning objective. They also diverge from human driving intuition, which prioritizes context directly relevant to the current decision. In contrast, \texttt{iPad} introduces two lightweight, proposal-centric auxiliary tasks: mapping and prediction, which are tightly coupled with the planning process. For each proposal, the mapping task predicts whether its states lie on-road or on-route, while the prediction task forecasts the future states of both the first object that will collide and the first object that is likely to collide (based on time-to-collision analysis) with the proposal planning trajectory.

Our main contributions are as follows:

1. \textbf{Iterative Proposal-Centric Paradigm}: We propose \texttt{iPad}, an end-to-end driving paradigm that centers the entire learning pipeline around sparse, learnable BEV proposals. \texttt{iPad} unifies feature extraction, mapping, prediction, and planning in a computationally efficient and interpretable manner.

2. \textbf{Proposal-Aware Feature Extraction}: We design ProFormer, a novel BEV encoder that integrates multi-view image features through proposal-anchored spatial attention. ProFormer jointly refines BEV queries and proposals, enabling high-quality multi-modal plan generation.

3. \textbf{Planning-Centric Auxiliary Tasks}: We introduce two lightweight, proposal-centric auxiliary tasks that enhance the planning process without introducing redundant computation or irrelevant scene modeling, improving both accuracy and efficiency.

4. \textbf{State-of-the-art performance}: \texttt{iPad} achieves state-of-the-art results on both the real-world NAVSIM~\cite{dauner2024navsim} and CARLA Bench2Drive~\cite{jia2024bench2drive} benchmarks. Notably, experiments show that \texttt{iPad} provides strong scalability and is over 10× more computationally efficient than UniAD~\cite{hu2023planning}.

\section{Related Work}

The goal of end-to-end (E2E) autonomous driving is to generate vehicle motion plans or control commands directly from raw sensor input, bypassing the need for task-specific modules such as detection and motion prediction. Early works such as ALVINN~\cite{pomerleau1988alvinn}, PilotNet~\cite{bojarski2016end}, and CIL~\cite{sauer2018conditional} leveraged large-scale human driving data to learn policies that directly map sensor observations to control actions. However, these models often suffered from poor interpretability and degraded performance due to issues like causal confusion~\cite{codevilla2019exploring}. To mitigate these limitations, recent research has explored incorporating intermediate representations, auxiliary tasks and proposal-based planning to enhance performance and robustness.

\textbf{Intermediate representations.} Two main categories of intermediate representations have been adopted in E2E autonomous driving: dense BEV grids and sparse query features. BEV representations naturally encode spatial relationships on the ground plane, making them ideal for joint perception and planning, and sensor fusion. ST-P3~\cite{hu2022st} was an early example that integrated detection, prediction, and planning into a unified BEV-based framework. Subsequent works—such as UniAD~\cite{hu2023planning}, VAD~\cite{jiang2023vad}, GenAD~\cite{zheng2024genad}, and GraphAD~\cite{zhang2024graphad}—follow a similar paradigm: generating dense BEV grid features from images and sequentially performing perception, prediction, and planning. Although effective, these methods are computationally expensive due to the high resolution required for accurate perception. To improve efficiency, a sparse query-centric paradigm has emerged, as seen in SparseDrive~\cite{sun2024sparsedrive}, DiFSD~\cite{su2024difsd}, and DriveTransformer~\cite{jia2025drivetransformer}. These methods use a limited number of learned queries to directly aggregate multi-view image features, avoiding costly view transformations. While this approach improves efficiency, it can still suffer from redundant computation and degraded planning performance due to excessive interactions with irrelevant agents—leading to causal confusion. Moreover, these approaches often overlook valuable prior knowledge (e.g., view transformations), resulting in suboptimal performance~\cite{li2024ego,zhai2023ADMLP}. However, all previous works typically build intermediate representations without explicit planning awareness, treating planning as a downstream task. In contrast, \texttt{iPad} integrates planning directly into the learning of intermediate representations via iterative proposal refinement. This joint optimization enables \texttt{iPad} to achieve both computational efficiency and high planning performance by focusing on planning-relevant features.

\textbf{Auxiliary E2E tasks.} To support the learning of interpretable intermediate representations, E2E methods often include auxiliary tasks such as object detection~\cite{sun2024sparsedrive}, BEV semantic segmentation~\cite{chitta2022transfuser}, occupancy prediction~\cite{hu2023planning}, and motion forecasting~\cite{jiang2023vad}. However, these tasks typically require high-resolution inputs and large models~\cite{li2022bevformer}, increasing computational cost. Furthermore, they often diverge from the core decision-making process of human drivers, who selectively focus on elements relevant to the current driving decision-making. To address these issues, we propose two lightweight, proposal-centric auxiliary tasks—mapping and prediction—that focus explicitly on modeling objects relevant to the ego vehicle’s planning proposals.

\textbf{Multi-modal planning.} Planning in autonomous driving is inherently multi-modal due to uncertainties in dynamic environments. However, most existing E2E methods~\cite{chitta2022transfuser,li2024enhancing,weng2024paradrive,li2024ego,wang2024driving} generate deterministic plans, which can lead to unrealistic or suboptimal behaviors. Recent works such as VADv2~\cite{chen2024vadv2} and Hydra-MDP~\cite{li2024hydra} address this by scoring a large set of fixed anchor trajectories to approximate the planning distribution. In contrast, SparseDrive~\cite{sun2024sparsedrive} predicts a small number of planning proposals in the final stage. However, fixed anchor vocabularies and limited proposal sets constrain expressiveness and adaptability. In contrast, \texttt{iPad} iteratively predicts and refines a dynamic set of planning proposals and leverages these proposals to guide feature extraction. This tight integration of planning and representation learning allows \texttt{iPad} to generate diverse, high-quality trajectories while maintaining efficiency.

\section{Method}

\begin{figure*}[t]
    \centering
    \includegraphics[width=\linewidth]{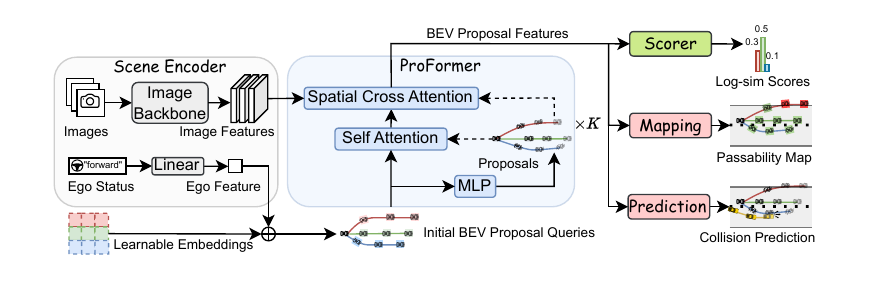}
    \vspace{-0.5cm}
    \caption{
    \textbf{Overview of the \texttt{iPad} framework}, consisting of four key components: the \textit{Scene Encoder} (gray) extracts image and ego features; the \textit{ProFormer} (blue) initializes BEV proposal queries with ego features and iteratively refines them using the image features; \textit{Scorer} (green) predicts a score for each proposal trajectory; and the \textit{Proposal-Centric Mapping and Prediction} (red) predict passability maps and agent future states related to potential collisions.}
    \vspace{-0.5cm}
\label{fig:ipad}
\end{figure*}

The overall framework of our \texttt{iPad} method is illustrated in~\cref{fig:ipad}. \texttt{iPad} comprises four components: \textbf{Scene Encoder} processes multi-view input images and ego vehicle status to extract both image and ego features; \textbf{ProFormer} iteratively refines trajectory proposals and queries with the extracted image features; \textbf{Scorer} predicts the planning performance of all final proposals and selects the one with the highest score as the output plan; \textbf{Proposal-Centric Mapping and Prediction} module predicts passability and collision risk for all final proposals during training, improving both interpretability and overall performance.

\subsection{Scene Encoder}

Our method takes two types of input: multi-view images and ego status. The multi-view images are processed through an image encoder, comprising a backbone network (e.g., ResNet-34~\cite{he2016deep}) and a neck, to extract multi-view image feature maps $\mI \in \sR^{I \times C \times H \times W}$, where $I$ is the number of image views, $C$ the feature channel dimension, $H$ the height, and $W$ the width of the feature maps. The ego status, including features such as ego current velocity, acceleration, and future commands, is encoded into the ego feature $\mE \in \sR^{1 \times C}$ using a linear layer.

\subsection{ProFormer}

We propose \textbf{ProFormer}, a proposal-centric BEV encoder built upon BEVFormer~\cite{li2022bevformer}, which iteratively refines BEV proposal queries by leveraging multi-view image features. ProFormer enhances the initial BEV queries by incorporating ego features. Moreover, unlike BEVFormer—which relies on a fixed dense grid of anchors to compute BEV features, leading to high computational overhead and limited planning awareness—ProFormer employs a learnable, proposal-based anchoring strategy that significantly improves both computational efficiency and planning relevance.

At each iteration $k = 0, \dots, K-1$, we first predict proposals $\mP_k \in \mathbb{R}^{N \times T \times 3}$ from current BEV proposal queries $\mQ_k \in \mathbb{R}^{N \times T \times C}$ using a MLP, where $N$ is the number of proposals and each proposal is a sequence of $T$ future states $(x, y, heading)$. We initialize proposal queries $\mQ_0$ by adding ego features $\mE$ to learnable positional embeddings. Then, we apply proposal-anchored deformable self-attention (SA) over the queries to capture temporal dependencies and interactions among proposals, using the predicted proposal positions as anchor points:
\begin{align}
    \text{SA}(\mQ_k^{n,t}, \mQ_k) = \text{DeformAttn}(\mQ_k^{n,t}, \mP_k^{n,t}(x,y), \mQ_k),
\end{align}
where $\mQ_k^{n,t} \in \mathbb{R}^{C}$ denotes the BEV query for the $n$-th proposal at time step $t$, and $\mP_k^{n,t}(x,y) \in \mathbb{R}^{2}$ is its predicted 2D position. The deformable attention mechanism~\cite{zhu2020deformable}, described in detail in~\cref{sec:deform}, computes attention by sampling a small set of points around each anchor, resulting in high efficiency. 

Following self-attention, we apply proposal-anchored deformable spatial cross-attention (SCA) to aggregate multi-view image features $\mI$, using the predicted four corner points of each proposal as anchors to better account for vehicle size and planning heading:
\begin{align}
    \text{SCA}(\mQ_k^{n,t} , \mI) &= \frac{1}{|\mathcal{V}_\text{hit}|} 
\sum_{i\in \mathcal{V}_\text{hit}}  \sum_{j}^{4} \sum_{z=1}^{N_\text{ref}}
    \text{DeformAttn}(\mQ_k^{n,t}, \mathcal{P}(\mP_k^{n,t},i,j,z), \mI_i),
\end{align}

where $\mI_i$ denotes the features from the $i$-th camera view. For each BEV query $\mQ_k^{n,t}$, each proposal’s four corner points are lifted into 3D pillars and sample $N_\text{ref}$ reference points per pillar. A projection function $\mathcal{P}$ maps the $z$-th reference point of the $j$-th corner onto the image plane of the $i$-th view. Since not all projected points fall within every view, we define the set of camera views that contain valid projections as $\mathcal{V}_\text{hit}$. Finally, a linear layer updates the refined proposal queries, producing $\mQ_{k+1}$ for the next iteration.

Notably, following previous auto-regressive methods such as GPT~\cite{achiam2023gpt} and diffusion models~\cite{ho2020denoising}, we design the ProFormer to \textbf{share weights} across iterations. To supervise proposal prediction at each iteration, we adopt a simple Minimum over N (MoN) loss~\cite{gupta2018social}, defined as:
\begin{align}
    \Ls_{proposal} =\sum_{k=0}^{K-1} \lambda^{K-1-k} \min_{n=1,\dots,N} \left\|\mP_k^n - \hat{\mP} \right\|_1,
\end{align}
where $\mP_k^n$ is the $n$-th proposal generated at iteration $k$, $\hat{\mP} \in \mathbb{R}^{T \times 3}$ is the expert trajectory, and $\lambda \in (0,1)$ is a discount factor that gradually relaxes the loss constraint for earlier iterations.

\subsection{Scorer}

To select a proposal as the planning, we learn a scorer to evaluate the final proposals $\mP_K$. The proposal with the highest predicted score is selected as the final planning trajectory. Specifically, we apply max pooling over the temporal dimension of BEV proposal features (\ie the final BEV proposal queries $\mQ_K \in \sR^{N \times T \times C}$), which are then fed into a multi-layer perceptron (MLP) to predict the scores $\mS \in \sR^{N \times 1}$. The score learning uses the binary cross-entropy (BCE) loss as: 
\begin{align}
    \Ls_{score}=\operatorname{BCE}(\mS, \hat{\mS}) ,
\end{align}
where $\operatorname{BCE}(x, y) = -y \log x + (1 - y) \log (1 - x)$. Considering the safety, efficiency, comfort of each proposal, we compute the ground-truth score following NAVSIM~\cite{dauner2024navsim}:
\begin{align}
    \hat{\mS} = \textit{NC} \times \textit{DAC} \times \frac{5\times \textit{EP}+ 5 \times \textit{TTC}+2\times \textit{Comf}}{12},
\label{eq:PDMS}
\end{align}
No at-fault Collision (NC), Drivable Area Compliance (DAC), Ego Progress (EP), Time-to-Collision (TTC), and Comfort (Comf) are sub-metrics obtained via a log-replay simulator. In this simulator, a controller is applied to recursively track the final proposal while other agents follow their recorded trajectory. For more details on obtaining the ground-truth sub-metrics, please refer to the~\cref{sec:score}. 

\subsection{Proposal-Centric Mapping and Prediction}

To enhance planning performance and interpretability, we design two light-weight plan-oriented auxiliary tasks: proposal-centric mapping and prediction. Unlike conventional auxiliary tasks that aim to model all objects in the scene, our approach focuses solely on predicting map and agent information relevant to each proposal. Moreover, since different proposals may lead to different predicted states for the same object, our method can also reflect perception and prediction uncertainty.

For \textbf{proposal-centric mapping}, we predict the on-road and on-route probabilities $\mM \in \mathbb{R}^{N \times T \times 2}$ for all proposals' simulated states using the BEV proposal features $\mQ_K$ as input to a MLP. The mapping task is trained by minimizing the BCE loss between the predicted probabilities and the ground-truth labels $\hat{\mM} \in \mathbb{R}^{N \times T \times 2}$:
\begin{align} 
\Ls_{map}=\operatorname{BCE}(\mM, \hat{\mM}).
\end{align} 
For \textbf{proposal-centric prediction}, we predict the future states of the first at-fault and likely-to-collide (with a time-to-collision below a defined threshold) agents, identified via the log-replay simulation. The agent state predictions are generated using a MLP applied to the max-pooled BEV proposal features $\mQ_K$. The predicted states $\mA \in \mathbb{R}^{N \times T \times 2 \times 9}$ include the 2D positions of the four corners $\mA_c\in \mathbb{R}^{N \times T \times 2 \times 4 \times 2}$, with corresponding validity labels $\mA_v \in \mathbb{R}^{N \times T \times 2 \times 1}$. The prediction task is supervised using an $\mathcal{L}_1$ loss on corner positions and a BCE loss on the validity labels:
\begin{align} 
\Ls_{pred} = \|\mA_c - \hat{\mA_c}\|_1+ w_{bce} \operatorname{BCE}(\mA_v, \hat{\mA_v}), 
\end{align} 
where $w_{bce}$ is the weight for the BCE term, and $\hat{\mA_c}$, $\hat{\mA_v}$ are the ground-truth corner positions and validity labels of the first at-fault and likely-to-collide agents.

\subsection{Training}

\texttt{iPad} can be end-to-end trained and optimized in a fully
differentiable manner. The overall loss function can be formulated as follows:
\begin{align}
    \Ls=\Ls_{proposal}+w_{score}\Ls_{score}+w_{map}\Ls_{map} +w_{pred}\Ls_{pred},
\end{align}
where $w_{score}$, $w_{map}$, and $w_{pred}$ are the weights for the scoring, mapping, and prediction losses, respectively. For more details on model structure, please refer to the~\cref{sec:model}.

\section{Experiments}

To evaluate the performance of our proposed method, we conducted experiments on both real-world open-loop and simulated closed-loop benchmarks.

\subsection{Open-Loop NAVSIM Benchmark}

For open-loop evaluations, we utilized the NAVSIM~\cite{dauner2024navsim} benchmark, which is based on real-world driving data. Unlike the popular nuScenes~\cite{caesar2020nuscenes} benchmark, which includes approximately 75\% of scenarios involving trivial straight driving, NAVSIM focuses on more complex driving situations. This simplicity in nuScenes allows methods like AD-MLP, which bypass perception entirely, to perform exceptionally well~\cite{zhai2023ADMLP}. Additionally, nuScenes primarily relies on simple displacement error and collision rate metrics, which fail to adequately capture real-world closed-loop driving performance, such as penalties for off-road driving.

\textbf{Dataset:} The NAVSIM dataset builds on the real-world nuPlan~\cite{Caesar2021nuplan} dataset, incorporating only relevant annotations and sensor data sampled at 2 Hz. It emphasizes scenarios involving intention changes where the ego vehicle's historical data cannot be extrapolated into a future plan. We trained and evaluated our model using the official \texttt{navtrain} and \texttt{navtest} splits, which contain 103k and 12k samples, respectively.

\textbf{Metrics:} The NAVSIM introduces a series of closed-loop metrics designed to evaluate open-loop simulation and reflect real-world closed-loop performance. The sub-metric scores align with our training sub-metric scores, with the addition of a PDM score (PDMS), defined as:
\begin{align}
    \textit{PDMS}= \textit{NC} \times \textit{DAC} \times \frac{5\times \textit{EP}+ 5 \times \textit{TTC}+2\times \textit{C}}{12},
\end{align}
where sub-metrics are derived from a non-reactive simulation over a 4-second horizon. A kinematic bicycle model, controlled by an LQR controller, tracks the planned trajectory to simulate the ego vehicle's movement at 10 Hz. These sub-metrics are computed based on the simulated trajectory, recorded trajectories of other agents, and map data.

\textbf{Results:} As shown in~\cref{tab:navsim}, our method significantly outperforms prior works on this benchmark in all metrics without relying on lidar input. The high driving area compliance underscores the effectiveness of our approach in extracting and utilizing planning-relevant map information. Furthermore, the superior ego progress highlights the expressiveness of our multi-modal planning framework.

\begin{table*}[t]
\caption{\textbf{Open-loop Results with Closed-loop Metrics on NAVSIM Benchmark.}}
\centering
\resizebox{1.0\textwidth}{!}
{
 \begin{tabular}{l|cc|ccccc|>{\columncolor[gray]{0.9}}c}
    \toprule
    Method & Input & Img. Backbone & NC $\uparrow$ &DAC $\uparrow$ & TTC $\uparrow$& Comf. $\uparrow$ & EP $\uparrow$ &  \textbf{PDMS} $\uparrow$  \\
    \midrule
    PDM-Closed~\cite{dauner2023parting} (Rule-based) & Perception GT & - & 94.6 & 99.8 & 86.9 & 99.9 & 89.9  & 89.1 \\
    \midrule
    VADv2-$\mathcal{V}_{8192}$~\cite{chen2024vadv2} & Camera \& Lidar & ResNet-34~\cite{he2016deep} & 97.2 & 89.1 & 91.6 & \textbf{100} & 76.0 & 80.9 \\
    Transfuser~\cite{chitta2022transfuser} & Camera \& Lidar & ResNet-34~\cite{he2016deep}  & 97.7 & 92.8 & 92.8 & \textbf{100} & 79.2 & 84.0 \\
    DRAMA~\cite{yuan2024drama} & Camera \& Lidar & ResNet-34~\cite{he2016deep} & 98.0 & 93.1 & \underline{94.8} & \textbf{100} & 80.1 & 85.5 \\
    Hydra-MDP-$\mathcal{V}_{8192}$-W-EP~\cite{li2024hydra} & Camera \& Lidar & ResNet-34~\cite{he2016deep} & \underline{98.3} &96.0 &94.6&\textbf{100}& 78.7& 86.5 \\
    DiffusionDrive~\cite{liao2024diffusiondrive} & Camera \& Lidar & ResNet-34~\cite{he2016deep} & 98.2  & \underline{96.2}  & 94.7  & \textbf{100}  & \underline{82.2}  & \underline{88.1}\\
    UniAD~\cite{hu2023planning} & Camera & ResNet-34~\cite{he2016deep}  & 97.8 & 91.9 & 92.9 & \textbf{100} & 78.8 & 83.4 \\
    LTF~\cite{chitta2022transfuser} & Camera & ResNet-34~\cite{he2016deep} & 97.4 & 92.8 & 92.4 & \textbf{100} & 79.0 & 83.8 \\
    PARA-Drive~\cite{weng2024paradrive} & Camera & ResNet-34~\cite{he2016deep}  & 97.9 & 92.4 & 93.0 & 99.8 & 79.3 & 84.0 \\
    \midrule
    \textbf{\texttt{iPad} (Ours)} & Camera & ResNet-34~\cite{he2016deep}  & \textbf{98.6} & \textbf{98.3} & \textbf{94.9} &  \textbf{100} &  \textbf{88.0} & \textbf{91.7} \\
    \bottomrule
\end{tabular}}
\vspace{-0.5cm}
\label{tab:navsim}
\end{table*}

\subsection{Closed-Loop Bench2Drive Benchmark}

Evaluating closed-loop driving performance in real-world scenarios is challenging, so we used the CARLA~\cite{dosovitskiy2017carla} simulator, employing the Bench2Drive benchmarks~\cite{jia2024bench2drive}.

\textbf{Dataset:} Bench2Drive provides a training dataset collected by the state-of-the-art expert model Think2Drive~\cite{li2024think2drive}. For fair comparisons, we utilized the \texttt{base} subset, which consists of 1,000 clips, with 950 clips allocated for training and 50 clips reserved for open-loop evaluation.

\textbf{Metrics:} 
Bench2Drive evaluates open-loop performance using the average $\Ls_2$ distance between the planned and expert trajectories over 2 seconds at 2 Hz. Closed-loop evaluations are conducted on 220 routes (approximately 150 meters each) across all CARLA towns, with each route featuring a safety-critical scenario. A PID controller tracks the planned trajectory at 20 Hz. Bench2Drive defines four closed-loop metrics:
\begin{itemize}
    \item Success Rate: The proportion of successfully completed routes within the allowed time and without traffic violations.
    \item Driving Score: The product of the route completion ratio and penalties for infractions, averaged across all routes.
    \item Efficiency: The ego vehicle’s average speed as a percentage of the average speed of nearby vehicles over 20 checkpoints along a route.
    \item Comfortness: The ratio of smooth trajectory segments to total segments. A trajectory segment is considered smooth if its lateral acceleration, yaw rate, yaw acceleration, and jerk remain within predefined thresholds.
\end{itemize}
Additionally, Bench2Drive evaluates five driving skills: merging, overtaking, emergency braking, yielding, and traffic sign adherence. The ability score for each skill is defined as the average success rate across all corresponding scenarios.

\begin{table*}[t]
\caption{\textbf{Open-loop and Closed-loop Results of E2E Methods on Bench2Drive Benchmark.}}
\centering
\small
\resizebox{1.0\textwidth}{!}
{\begin{tabular}{l|c|c|ccc>{\columncolor[gray]{0.9}}c}
\toprule
\multirow{2}{*}{\textbf{Method}}  & \multirow{2}{*}{\textbf{Latency}} & \textbf{Open-loop} & \multicolumn{4}{c}{\textbf{Closed-loop}} \\ \cmidrule{3-7} 
& &  Avg. L2 $\downarrow$  & Efficiency $\uparrow$ & Comfortness $\uparrow$  & Success Rate (\%) $\uparrow$ & \textbf{Driving Score} $\uparrow$  \\ 
 \midrule
AD-MLP~\cite{zhai2023ADMLP}  & \textbf{4 ms}  & 3.64      & 48.45 & 22.63  &  0.00 &  18.05  \\ 
UniAD-Tiny~\cite{hu2023planning} & 445 ms & 0.80  & 123.92 & \underline{47.04} & 13.18  &   40.73 \\
UniAD-Base~\cite{hu2023planning} &  558 ms & \underline{0.73} & 129.21 & 43.58  & 16.36 & 45.81\\
VAD~\cite{jiang2023vad}  & 359 ms   & 0.91  & \underline{157.94} & 46.01  & 15.00 &  42.35  \\ 
DriveTransformer~\cite{jia2025drivetransformer} & 212 ms & \textbf{0.62}& 100.64 & 20.78  & \underline{35.01}   &  63.46  \\
\textbf{\texttt{iPad} (Ours)} & \underline{43 ms} & 0.97 & \textbf{161.31}  & 28.21  &    \textbf{35.91} & \textbf{65.02} \\
\midrule
TCP*~\cite{wu2022trajectory}  & 71 ms  & 1.70    & 54.26 & 47.80   & 15.00 &  40.70  \\ 
TCP-ctrl*   &  71 ms    & -    & 55.97 & \textbf{51.51}   & 
7.27 &  30.47   \\
TCP-traj*   & 71 ms    & 1.70   & 76.54 & 18.08  & 30.00   & 59.90 \\
TCP-traj w/o distillation   & 71 ms  & 1.96        & 78.78 & 22.96 & 20.45  & 49.30 \\
ThinkTwice*~\cite{jia2023thinktwice} & 762 ms  & 0.95   & 69.33 & 16.22 & 31.23  & 62.44\\
DriveAdapter*~\cite{jia2023driveadapter} & 931 ms &  1.01 & 70.22 & 16.01 & 33.08  &  64.22 \\ \bottomrule
\end{tabular}}
\vskip 5pt
\footnotesize{
$^*$ denotes expert feature distillation. All latencies are measured as the average inference time (including input preparation, model inference, and control generation) during CARLA evaluation on NVIDIA RTX 4090 GPU except for DriveTransformer, ThinkTwice and DriveAdapter on A6000 from~\cite{jia2025drivetransformer}.}
\label{tab:bench2drive}
\vspace{-0.4cm}
\end{table*}

\begin{table*}[t]
\caption{\textbf{Multi-Ability Results of E2E Methods on Bench2Drive Benchmark.}}
\centering
\small
\resizebox{1.0\textwidth}{!}
{
\begin{tabular}{l|ccccc|>{\columncolor[gray]{0.9}}c}
\toprule
\multirow{2}{*}{\textbf{Method}} & \multicolumn{5}{c}{\textbf{Ability} (\%) $\uparrow$}                        \\ \cmidrule{2-7} 
& \multicolumn{1}{c}{Merging} & \multicolumn{1}{c}{Overtaking} & \multicolumn{1}{c}{Emergency Brake} & \multicolumn{1}{c}{Give Way} & Traffic Sign & \textbf{Mean} \\ \midrule
AD-MLP~\cite{zhai2023ADMLP}        & 0.00 & 0.00   
& 0.00        & 0.00         &  4.35    & 0.87         \\
UniAD-Tiny~\cite{hu2023planning}  & 8.89        & 9.33   & 20.00       & 20.00        & 15.43    & 14.73        \\ 
UniAD-Base~\cite{hu2023planning}   & 14.10       & 17.78   & 21.67       &  10.00       & 14.21    & 15.55    \\ 
VAD~\cite{jiang2023vad}  & 8.11       & 24.44        
& 18.64       &  20.00       & 19.15    & 18.07       \\
DriveTransformer~\cite{jia2025drivetransformer} & 17.57 & \textbf{35.00} & 48.36 & 40.00   & 52.10 & 38.60\\
\textbf{\texttt{iPad} (Ours)} &  \textbf{30.00}   & 20.00 & \textbf{53.33} & \textbf{60.00}  & 49.47 & \textbf{42.56} \\
\midrule
TCP*~\cite{wu2022trajectory}  & 16.18        & 20.00      & 20.00        &  10.00         & 6.99     & 14.63  \\
TCP-ctrl*                           & 10.29        & 4.44 & 10.00        &  10.00          & 6.45     & 8.23     \\
TCP-traj*                     & 8.89       & 24.29        & \underline{51.67}    &  40.00       & 46.28    & 34.22\\ 
TCP-traj w/o distillation   & 17.14       & 6.67           & 40.00  &  \underline{50.00}         & 28.72    & 28.51  \\ 
ThinkTwice*~\cite{jia2023thinktwice}  & 27.38   &  18.42  
& 35.82   &  \underline{50.00}       & \underline{54.23}    & 37.17  \\ 
DriveAdapter*~\cite{jia2023driveadapter}  &\underline{28.82}  & \underline{26.38}          & 48.76     &  \underline{50.00}       & \textbf{56.43}  & \underline{42.08}        \\ \bottomrule
\end{tabular}}
\label{tab:bench2drive2}
\label{tab:ability}
\vspace{-0.5cm}
\end{table*}

\textbf{Results:}  As shown in~\cref{tab:bench2drive}, our method achieves state-of-the-art performance in success rate and driving score without relying on an expert model. Furthermore, our lightweight network design result in significantly reduced latency, making it highly efficient for real-time applications. As demonstrated in~\cref{tab:bench2drive2}, our method also achieve best average performance over five driving abilities, showcasing its versatility and robustness in handling diverse and challenging scenarios.

\subsection{Ablation Studies}

To evaluate the contributions of individual components, we conducted ablation studies using the NAVSIM benchmark. Comfort metrics were omitted, as all ablated models consistently achieved a perfect score of 100.

\begin{table*}[t]
\centering
\caption{\textbf{Ablation Studies on the NAVSIM Benchmark.}}
\centering
\resizebox{1.0\textwidth}{!}
{
 \begin{tabular}{cccc |cccc|>{\columncolor[gray]{0.9}}c}
\toprule
 Proposal Refinement & BEV Encoder &  Mapping Module &  Prediction Module   & NC $\uparrow$ &DAC $\uparrow$ & TTC $\uparrow$ & EP $\uparrow$ & \textbf{ PDMS $\uparrow$}  \\
\midrule
No & BEVFormer &  General~\cite{chitta2022transfuser} & General~\cite{chitta2022transfuser} & 97.6 &   93.0 & 92.9 & 68.9  &  78.5 \\
Yes &BEVFormer &  General~\cite{chitta2022transfuser} & General~\cite{chitta2022transfuser} & 96.9 &  93.2 &  90.8 & 71.5  &  79.4 \\
Yes & ProFormer  &  General~\cite{chitta2022transfuser} & General~\cite{chitta2022transfuser} & 98.1 & 96.5 & 94.3 & 84.2 & 89.8\\
Yes & ProFormer  &  Proposal-centric & General~\cite{chitta2022transfuser}  & 98.3 & 97.9 & 94.4 & 85.9 & 90.5 \\
Yes & ProFormer  & Proposal-centric & Proposal-centric & \textbf{98.6} & \textbf{98.3} & \textbf{94.9}  &  \textbf{88.0} & \textbf{91.7} \\
\bottomrule
\end{tabular}
}
\vspace{-0.3cm}
\label{tab:auxillary}
\end{table*}

\textbf{Effectiveness of proposal-centric BEV encoder:} We evaluate the effectiveness of our proposal-centric BEV encoder by replacing ProFormer with the baseline BEVFormer. First, to exclude the impact of the intermediate proposal learning, we conduct an experiment using BEVFormer to also predict proposals at each iteration. As shown in~\cref{tab:auxillary}, this naive approach to proposal learning yields limited gains, as the image feature extraction process in BEVFormer does not incorporate the predicted proposals. We then replace BEVFormer with our ProFormer, which leads to a significant improvement in all planning metrics—highlighting the benefit of our proposal-aware spatial cross-attention mechanism.

\textbf{Advantages of proposal-centric auxiliary tasks:} To evaluate the impact of our auxiliary task design, we substitute the standard mapping and prediction tasks from Transfuser~\cite{chitta2022transfuser} with our proposal-centric variants. As shown in~\cref{tab:auxillary}, replacing the proposal-centric mapping task results in a drop in driving area compliance. Similarly, replacing the proposal-centric prediction task degrades performance in terms of no at-fault collisions and time-to-collision. These results demonstrate the value of our planning-oriented auxiliary tasks in enhancing driving performance.

\subsection{Scalability}

\begin{figure}[h]
\begin{center}
\vspace{-0.3cm}
\includegraphics[width=\linewidth]{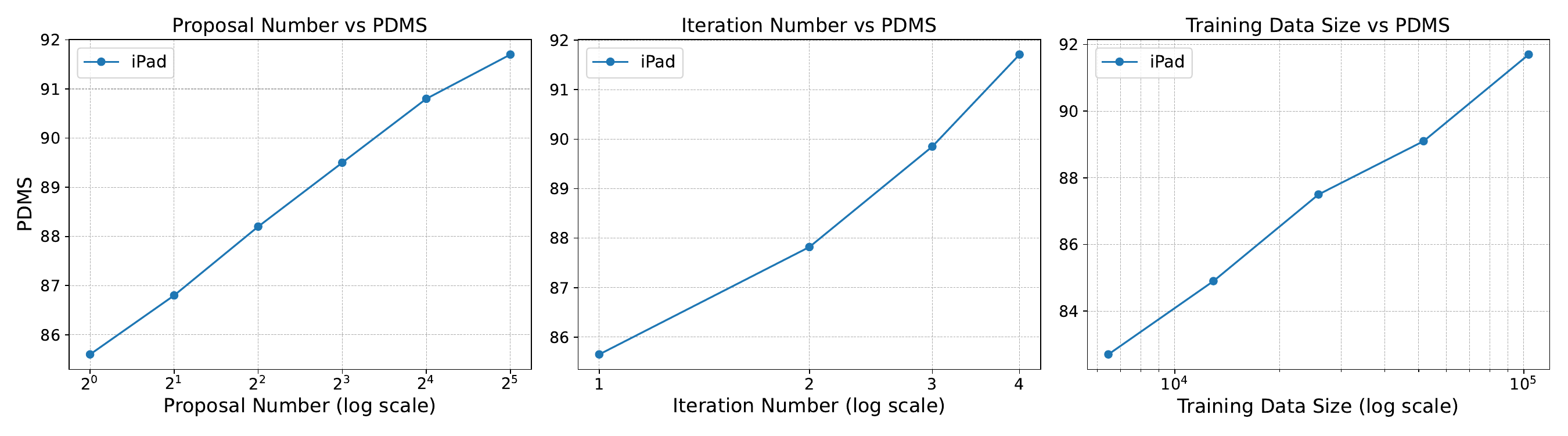}
\end{center}
\vspace{-0.3cm}
\caption{
\textbf{Scaling law in \texttt{iPad}.} The PDM score performance on the NAVSIM Benchmark increases logarithmically with the proposal number, iteration number and training data size,}
\vspace{-0.3cm}
\label{fig:scale}
\end{figure}

We investigate the trend in \texttt{iPad}'s planning performance as the proposal number, iteration number, and training data size increase. The final PDM score on the test set of the NAVSIM Benchmark is evaluated, and the results are presented in ~\cref{fig:scale}. A clear power-law scaling trend is observed for the PDM score with respect to the proposal number, iteration count, and training data size. Specifically, a higher number of proposals enhances the flexibility of the planning distribution and effectively expands the model’s representation capacity. More refinement iterations improve the accuracy of the proposals by leveraging a greater number of image features, while larger training data volumes contribute to better generalization of the model.

\subsection{Qualitative Analysis}

\begin{figure}[t]
\begin{center}
\includegraphics[width=\linewidth]{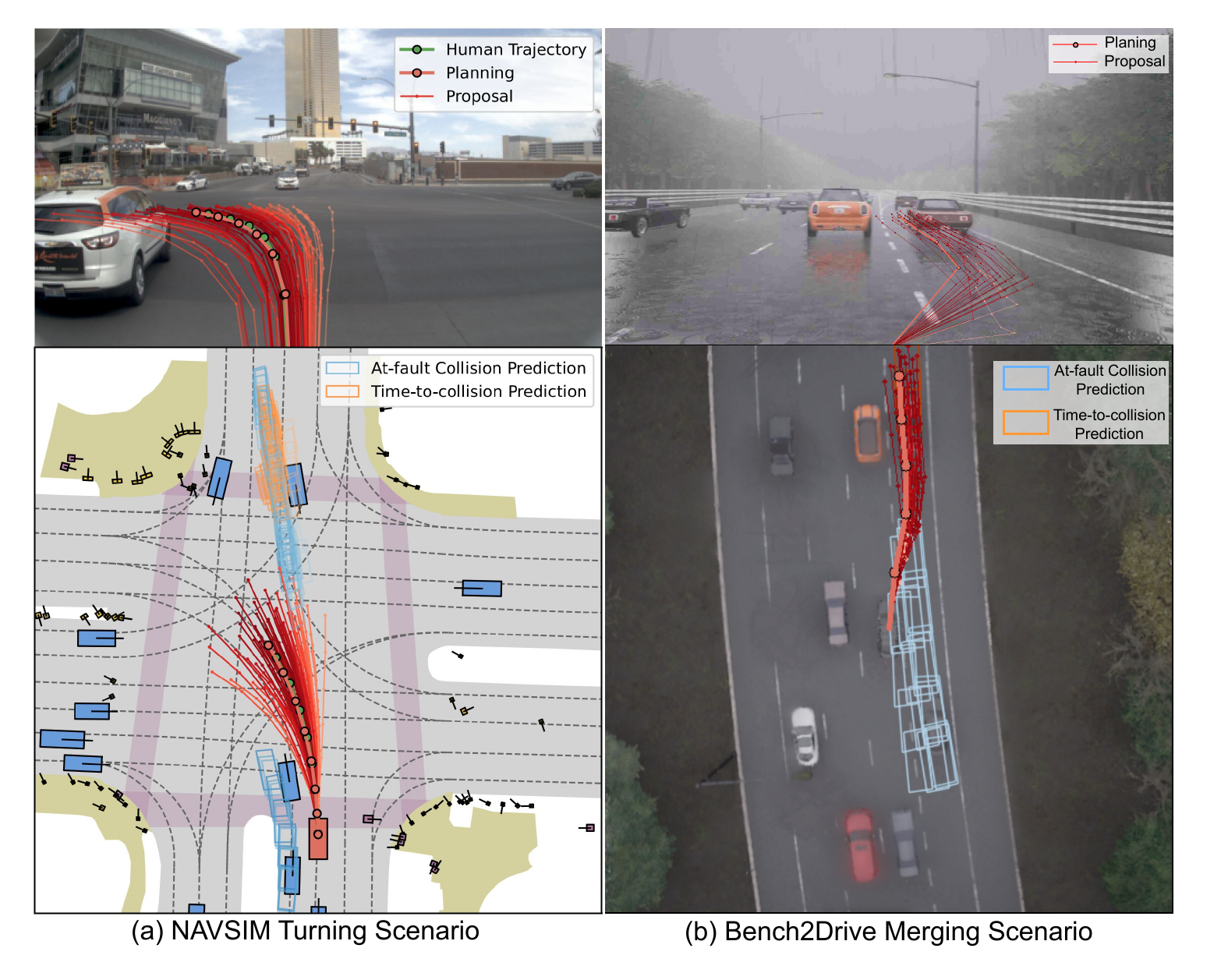}
\end{center}
\vspace{-0.3cm}
\caption{
\textbf{Qualitative planning and collision prediction results on NAVSIM and Bench2Drive.} Proposal lines are shaded with brightness proportional to their predicted scores, while the brightness of predicted agent boxes reflects their associated proposals.}
\vspace{-0.3cm}
\label{fig:navsim}
\end{figure}

We visualized the planning and prediction results of our method in NAVSIM and Bench2Drive scenarios. As illustrated in~\cref{fig:navsim}, in a NAVSIM turning scenario, our method generates diverse, human-like planning proposals closely aligned with actual human trajectories. The prediction results accurately reflect collision risks, prioritizing central proposals with higher scores.  In a Bench2Drive merging scenario, our method produced a collision-free planning, with predictions effectively highlighting collision risks and prioritizing conservative merging proposals. More qualitative examples can be found in \cref{sec:qualitative}.

\section{Limitations}
\label{sec:limitation}

Our work has two primary limitations. First, we do not incorporate historical image and status information to maintain efficiency. However, utilizing historical data could help address occlusion issues and enhance the accuracy of trajectory predictions for other agents. Second, we lack real-world closed-loop evaluations. While our open-loop evaluations use real-world data, closed-loop performance remains uncertain due to the distribution shift. Simulated closed-loop evaluations face challenges from the sim-to-real gap, as simulations cannot fully capture the complexity and unpredictability of real-world driving. Factors such as corner cases, unexpected human behavior, and diverse environmental conditions are often inadequately modeled.

\section{Conclusion} 
\label{sec:conclusion}
We presented \texttt{iPad}, a novel end-to-end autonomous driving framework that rethinks the role of planning in the E2E learning paradigm. By placing sparse, learnable proposals at the center of perception, prediction, and planning, \texttt{iPad} offers a unified, interpretable, and computationally efficient alternative to dense BEV grid-based methods. Our proposed ProFormer encoder and lightweight proposal-centric auxiliary tasks enable the model to focus on planning-relevant information while avoiding unnecessary computation and spurious correlations. Extensive experiments on challenging real-world and simulation benchmarks demonstrate that \texttt{iPad} achieves state-of-the-art performance while being significantly more efficient than prior work.



\bibliographystyle{plain}
\bibliography{main}

\newpage
\appendix
\section*{\Large{
\textit{Appendix}}}
\vskip8pt

\section{Detailed Mechanism in ProFormer}
\label{sec:deform}
\textbf{Deformable attention defintion:} The deformable attention is defined as:
\begin{align}
\text{DeformAttn}(q, p, x) = \sum_{i=1}^{N_\text{head}} \mathcal{W}_i\sum_{j=1}^{N_\text{key}} \mathcal{A}_{ij} \cdot \mathcal{W}'_i x(p+ \Delta p_{ij}),
\label{eq:single_deform_attn_fun}
\end{align}
where $q$, $p$, $x$ represent the query, reference point and input features, respectively. $i$ indexes the attention head, and $N_\text{head}$ denotes the total number of attention heads.
$j$ indexes the sampled keys, and $N_\text{key}$ is the total sampled key
number for each head.
$W_i\!\in\!\mathbb{R}^{C\times (C / H_{\rm head})}$ and $W'_i\!\in\!\mathbb{R}^{(C / H_{\rm head})\times C}$ are the learnable weights, where $C$ is the feature dimension. $A_{ij}\!\in\![0,1]$ is the predicted attention weight, and is normalized by $\sum_{j=1}^{N_\text{key}} A_{ij}\!= \!1$. $\Delta p_{ij}\!\in\!\mathbb{R}^2$ are the predicted offsets to the reference point $p$. $x(p + \Delta p_{ij})$ represents the feature at location $p + \Delta p_{ij}$, which is 
extracted by bilinear interpolation as in Dai~\etal\cite{dai2017deformable}.

\begin{figure}[h]
\begin{center}
\includegraphics[width=0.75\linewidth]{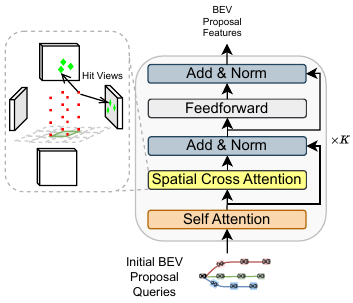}
\end{center}
\vspace{-0.3cm}
\caption{Detailed architecture of ProFormer. The proposals are used to query deformable proposal-centric image features $\mI$ (yellow) to update the proposal features.}
\vspace{-0.4cm}
\label{fig:ProFormer}
\end{figure}

\textbf{Spatial cross attention details:} Spatial cross-attention, shown in~\cref{fig:ProFormer}, computes the attention between proposal queries and the image features $I$ using the predicted proposal. For each proposal pose, the vehicle's four corner points are calculated as BEV anchor points, incorporating vehicle size and planned heading information. Reference points sampled from pillars lifted from these anchors are projected onto 2D image views, and image features around these projected points are aggregated using deformable attention. For one BEV
query, the projected 2D points can only fall on some views, and other views are not hit. Here, we term them the hit views.

\newpage

\section{Model Details}
\label{sec:model}
For both datasets, the same model architecture is used, whose hyper-parameters are listed in~\cref{tab:Hyper-parameters}.  All models are trained on a single NVIDIA H800 GPU with a batch size of 64 for 20 epochs, using the Adam optimizer with a learning rate of $1 \times 10^{-4}$. For efficiency, we only use downsampled images from the front, front-left, front-right, and back views as input. 

\begin{table}[h]
\caption{Hyper-parameters}
\vspace{-0.3cm}
\label{tab:Hyper-parameters}
\begin{center}
\begin{tabular}{lc}
\multicolumn{1}{c}{\bf Hyper-parameter}  &\multicolumn{1}{c}{\bf Value}
\\ \hline 
    Proposal number $N$ & 64 \\
    Iteration number $K$ & 4\\
    Planning time step interval & 0.5s \\
    Channel dimension $C$ & 256 \\
    Hidden size &  256 \\
    Feed-forward size &  1024 \\

    Pillar reference point number $N_{ref}$& 4 \\

    Proposal loss discount $\lambda$ & 0.1 \\
    Score loss weight $w_{score}$ & 1 \\
    Map loss weight $w_{map}$ & 2 \\
    Prediction loss weight $w_{pred}$ & 1 \\
    Prediction BCE loss weight $w_{bce}$ & 0.1 \\

    \midrule
    NAVSIM future planning horizon $T$ & 8 \\ 
    NAVSIM image input down-sample rate  & 0.4 \\

    \midrule
    Bench2Drive future planning horizon $T$ & 6  \\ 
    Bench2Drive image input down-sample rate  & 0.64 \\
    \bottomrule
\\
\end{tabular}
\end{center}
\vspace{-1.0cm}
\end{table}

\section{Training Scoring}
\label{sec:score}

To efficiently obtain ground-truth scores for the final proposals during training, we employ parallelized computation using \textbf{Ray} for multi-processing.

\subsection{NAVSIM Scoring}

For \textbf{NAVSIM}, we use the official log-replay simulator with an LQR controller operating at 10~Hz over a 4-second horizon. Final scores are derived based on the following official sub-metrics:

\begin{itemize}
    \item \textbf{No At-Fault Collision (NC)}: Set to 0 if, at any simulation step, the proposal's bounding box intersects with other road users (vehicles, pedestrians, or bicycles). Collisions that are not considered ``at-fault'' in the non-reactive environment (e.g., when the ego vehicle is stationary) are ignored. For collisions with static objects, a softer penalty of 0.5 is applied.
    
    \item \textbf{Drivable Area Compliance}: Set to 0 if, at any simulation step, any corner of the proposal state lies outside the drivable area polygons.
    
    \item \textbf{Time-to-Collision (TTC)}: Initialized to 1. Set to 0 if, at any point during the 4-second horizon, the ego vehicle’s projected time-to-collision---assuming constant velocity and heading---is less than 1 second.
    
    \item \textbf{Comfort}: Set to 0 if, at any simulation step, motion exceeds any of the following thresholds:
    \begin{itemize}
        \item Lateral acceleration $>$ 4.89 m/s$^2$
        \item Longitudinal acceleration $>$ 2.40 m/s$^2$
        \item Longitudinal deceleration $>$ 4.05 m/s$^2$
        \item Absolute jerk $>$ 8.37 m/s$^3$
        \item Longitudinal jerk $>$ 4.13 m/s$^3$
        \item Yaw rate $>$ 0.95 rad/s
        \item Yaw acceleration $>$ 1.93 rad/s$^2$
    \end{itemize}
    
    \item \textbf{Ego Progress}: Measures the agent’s progress along the route center, normalized by a safe upper bound estimated by the PDM-Closed planner. The final ratio is clipped to $[0, 1]$, and scores are discarded if the upper bound is below 5 meters or the progress is negative.
\end{itemize}

\subsection{Bench2Drive Scoring}

For \textbf{Bench2Drive}, we utilize a log-replay simulator with a perfect controller operating at 2~Hz over a 3-second horizon. Evaluation is based on the following sub-metrics:

\begin{itemize}
    \item \textbf{No Collision (NC)}: Set to 0 if, at any simulation step, the proposal's bounding box intersects with any object (vehicles, bicycles, pedestrians, traffic signs, traffic cones, or traffic lights).
    
    \item \textbf{Drivable Area Compliance (DAC)}: Set to 0 if, at any simulation step, any corner of the proposal state lies off-road or all centers off-route.
    
    \item \textbf{Time-to-Collision (TTC)}: Set to 0 if, at any point during the 3-second horizon, the ego vehicle’s projected time-to-collision is less than 1 second.
    
    \item \textbf{Comfort}: Set to 0 if the proposal’s acceleration or turning rate exceeds the expert trajectory's maximum values.
    
    \item \textbf{Ego Progress}: Defined as the ratio of the ego progress along the expert trajectory, conditioned on being collision-free and on-road. If the ratio exceeds 1, its reciprocal is taken.
\end{itemize}

\subsection{Relations between Open-loop and Closed-loop Scores}

\begin{figure*}[h]
\begin{center}
\includegraphics[width=0.8\linewidth]{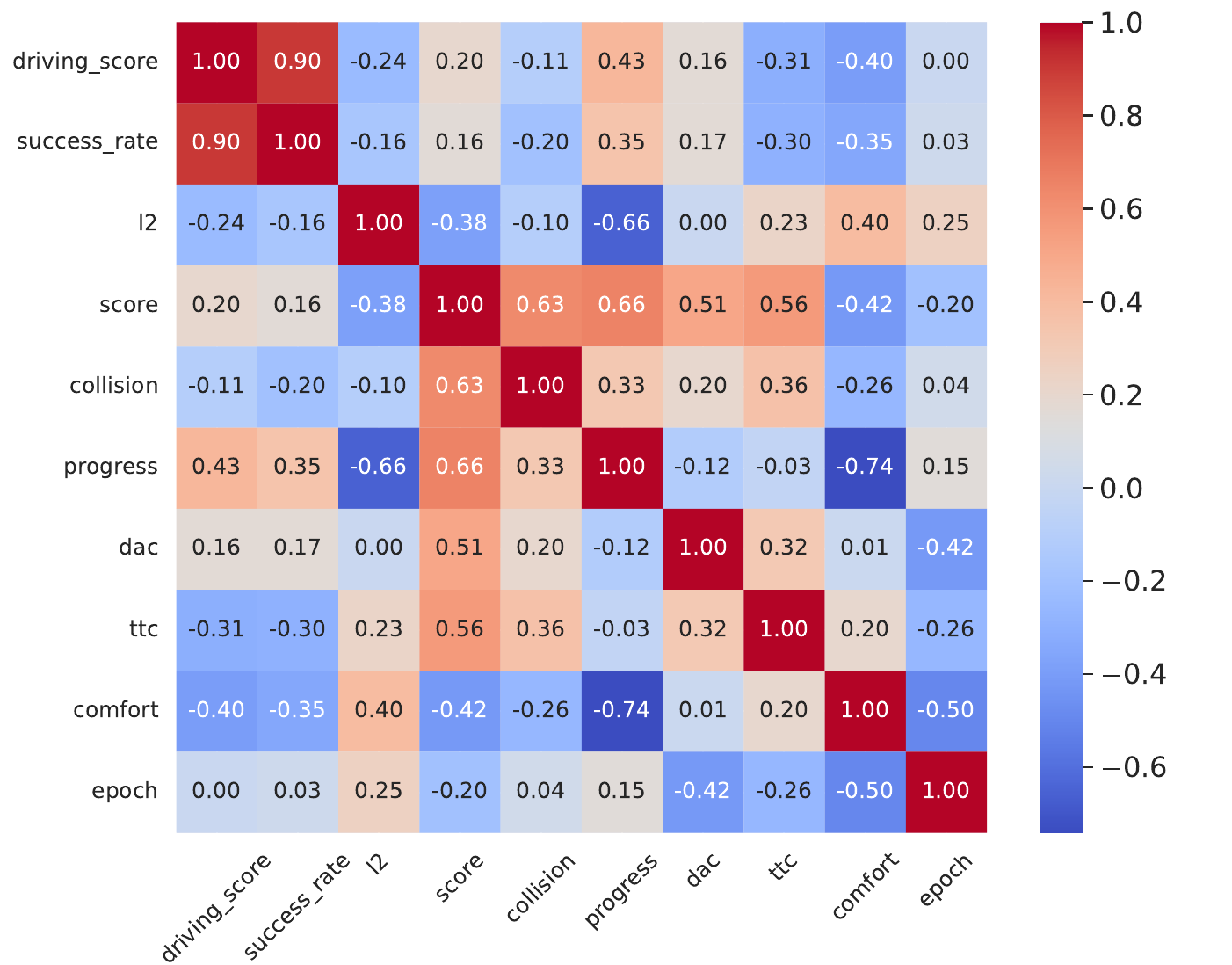}
\end{center}
\vspace{-0.4cm}
\caption{\textbf{Correlation Matrix of Open-loop and Closed-loop Driving Metrics}}
\vspace{-0.3cm}
\label{fig:correlation}
\end{figure*}

To evaluate the effectiveness of our scoring method, we analyze the relationship between open-loop validation metrics (L2, Score, NC, DAC, TTC, Progress, Comfort), closed-loop metrics (driving score, success rate), and training epoch. Specifically, we test 20 checkpoints—randomly sampled after the 10th training epoch, when the model has stabilized—for both the shared and non-shared versions of iPad. We compute the correlation coefficients between all metrics, as shown in~\cref{fig:correlation}.

Our results show that both the Score and Progress metrics are positively correlated with the final closed-loop driving performance. In contrast, the collision-related metrics (NC and TTC) exhibit a negative correlation with the closed-loop metrics, which may be attributed to a mismatch between agent behaviors: the real-world agents are reactive, while our log-sim assumes them to be non-reactive. Additionally, we observe a negative correlation between the open-loop L2 metric and closed-loop performance, consistent with findings in prior work~\cite{Guo2023CCIL}. Finally, the Comfort metric also shows a negative correlation with closed-loop driving scores, likely due to the high frequency of hazardous scenarios in the Bench2Drive benchmark that require abrupt braking.

\section{More Qualitative Results}
\label{sec:qualitative}
We show more qualitative results in both NAVSIM and Bench2Drive closed-loop testing scenarios. 

\subsection{Proposal Refinement}
\begin{figure*}[h]
\begin{center}
\includegraphics[width=\linewidth]{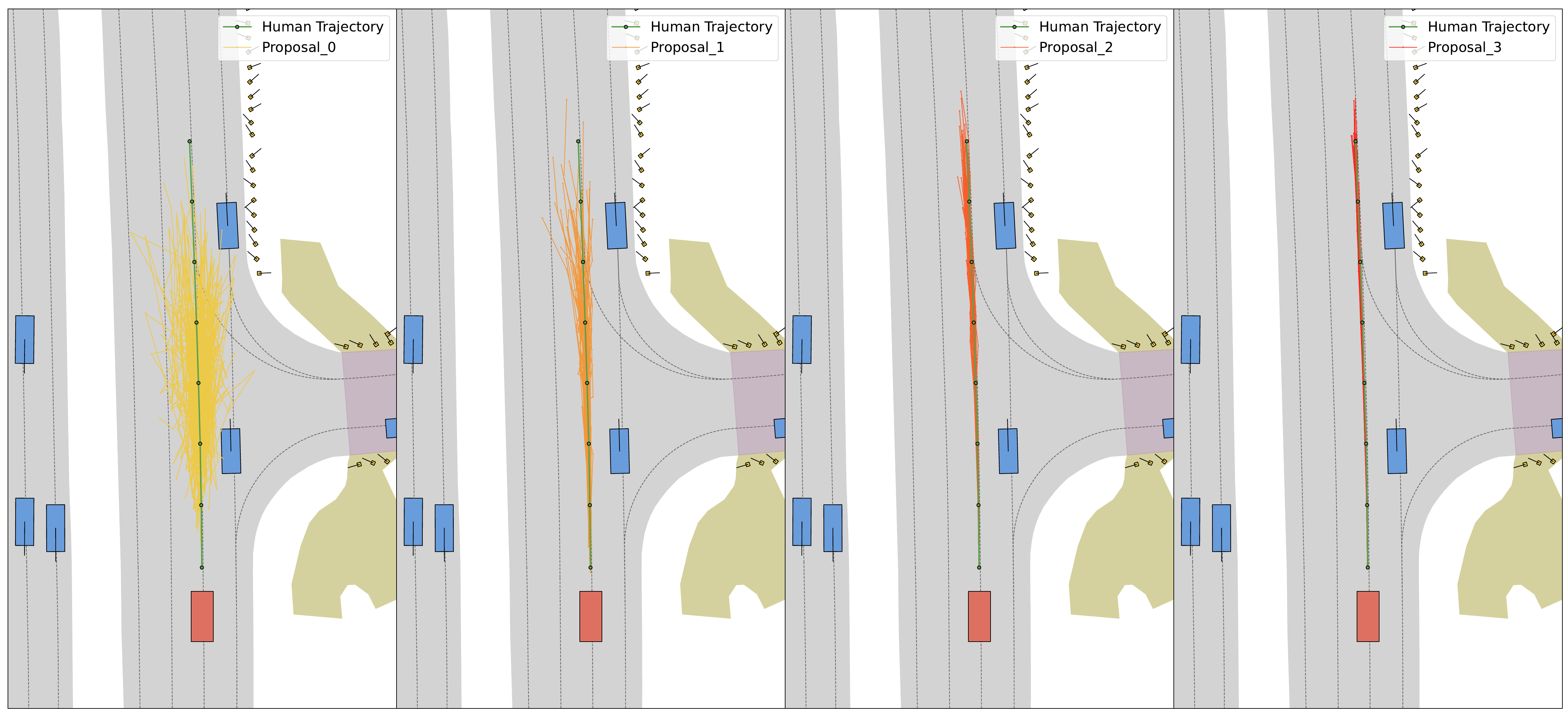}
\end{center}
\vspace{-0.3cm}
\caption{\textbf{Proposal prediction results at all iterations in a NAVSIM scenario.}  }
\vspace{0.3cm}
\label{fig:refine}
\end{figure*}

The~\cref{fig:refine} demonstrate that \texttt{iPad} can gradually refine the proposals, making it more similar to human trajectory. 

\begin{figure*}[h]
\begin{center}
\includegraphics[width=\linewidth]{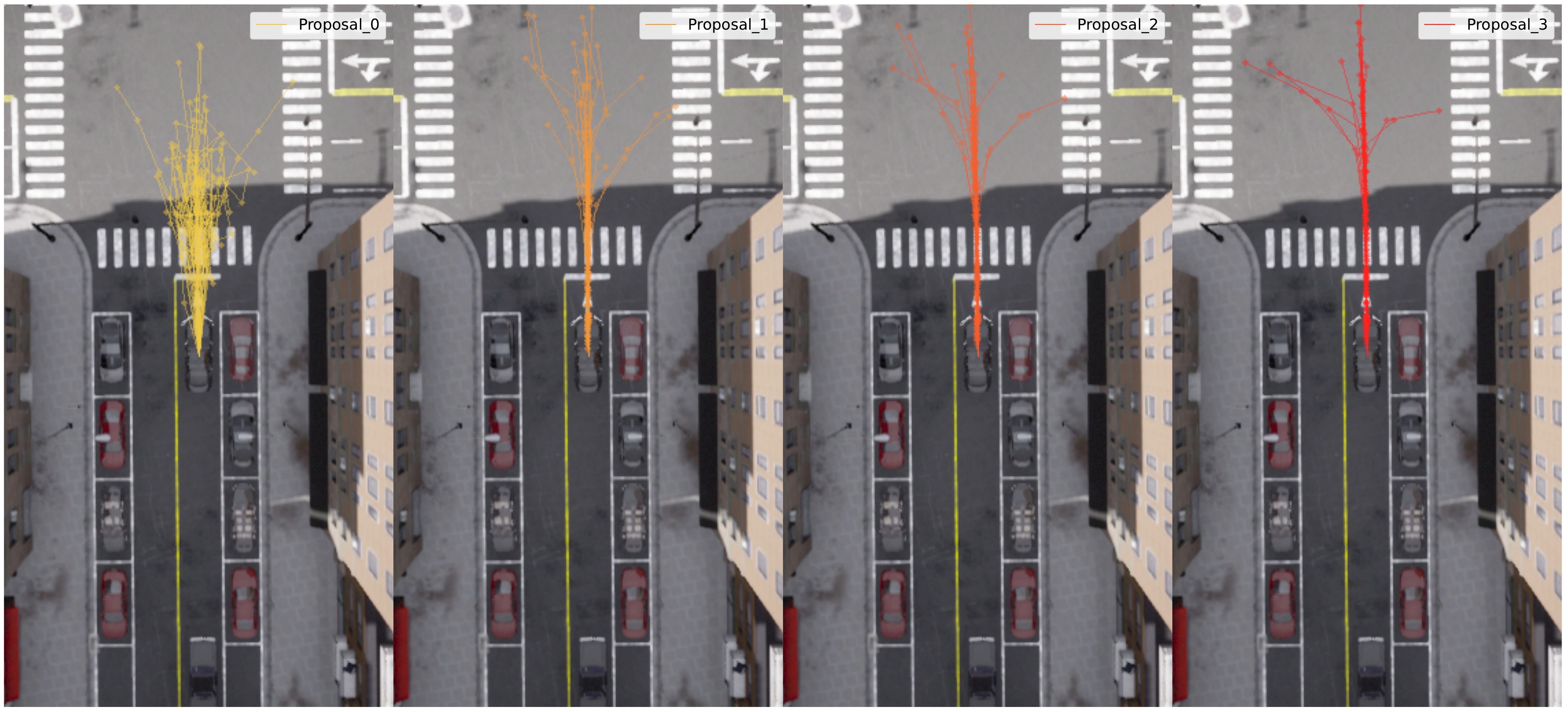}
\end{center}
\vspace{-0.3cm}
\caption{\textbf{Proposal prediction results at all iterations in a Bench2Drive scenario.}  }
\label{fig:refine1}
\end{figure*}

The~\cref{fig:refine1} demonstrate that \texttt{iPad} can gradually refine the proposals, while keeping the multi-modality in the intersection scenarios. 

\newpage

\subsection{Mapping}

\begin{figure*}[h]
\begin{center}
\includegraphics[width=\linewidth]{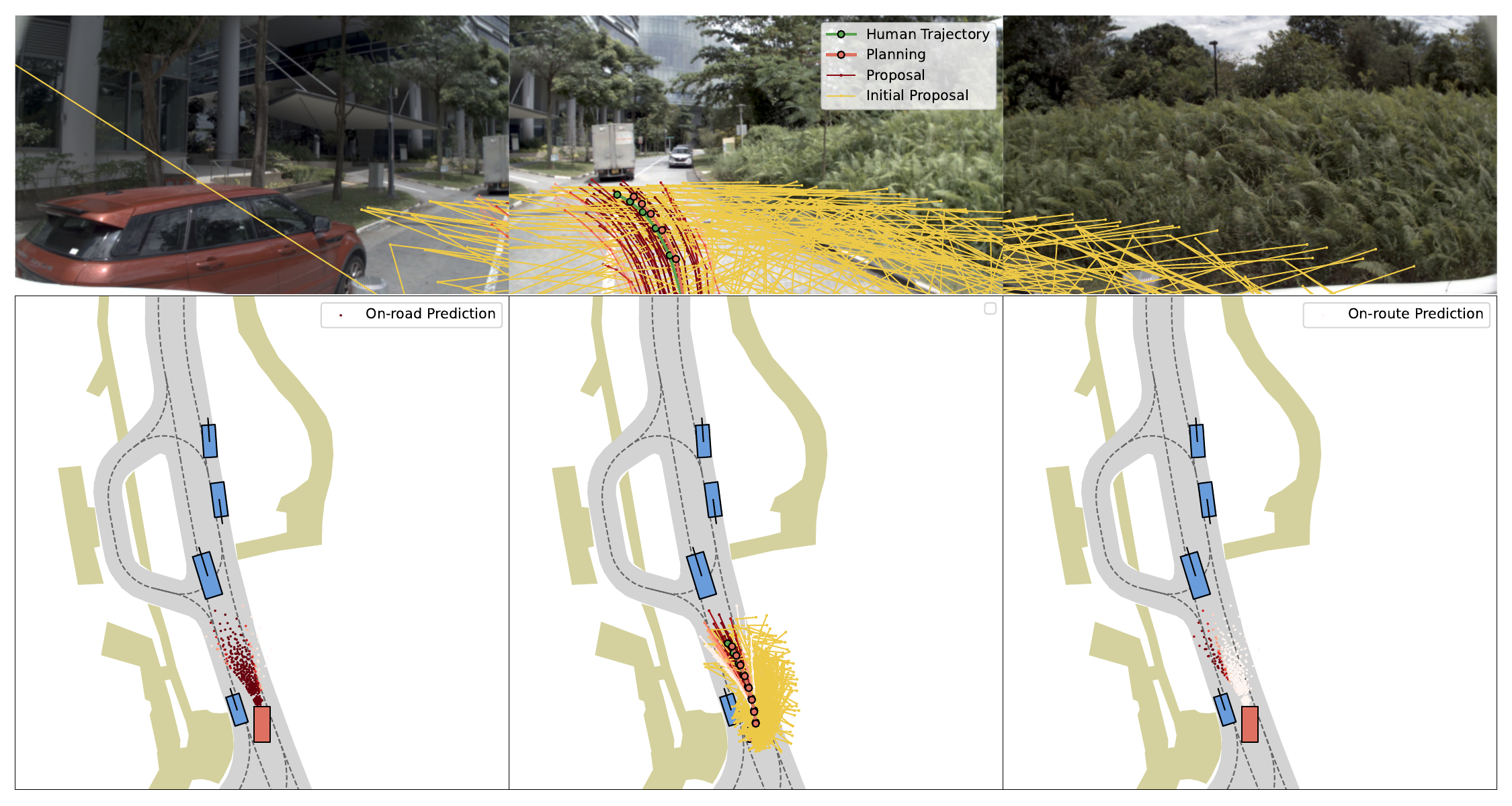}
\end{center}
\vspace{-0.3cm}
\caption{\textbf{Passability prediction results in a NAVSIM scenario.} The lightness of the proposal lines or points decreases with their scores or predicted on-road or on-route probabilities. The proposal state is off-road if any corner point is off-road.}
\vspace{0.3cm}
\label{fig:off_route}
\end{figure*}

The~\cref{fig:off_route} demonstrate that \texttt{iPad} can generate accurate on-road and on-route probability predictions in NAVSIM scenarios, being aware of the proposal heading and vehicle size. Therefore, a on-road and on-route proposal is chosen as the planning.

\begin{figure*}[h]
\begin{center}
\includegraphics[width=\linewidth]{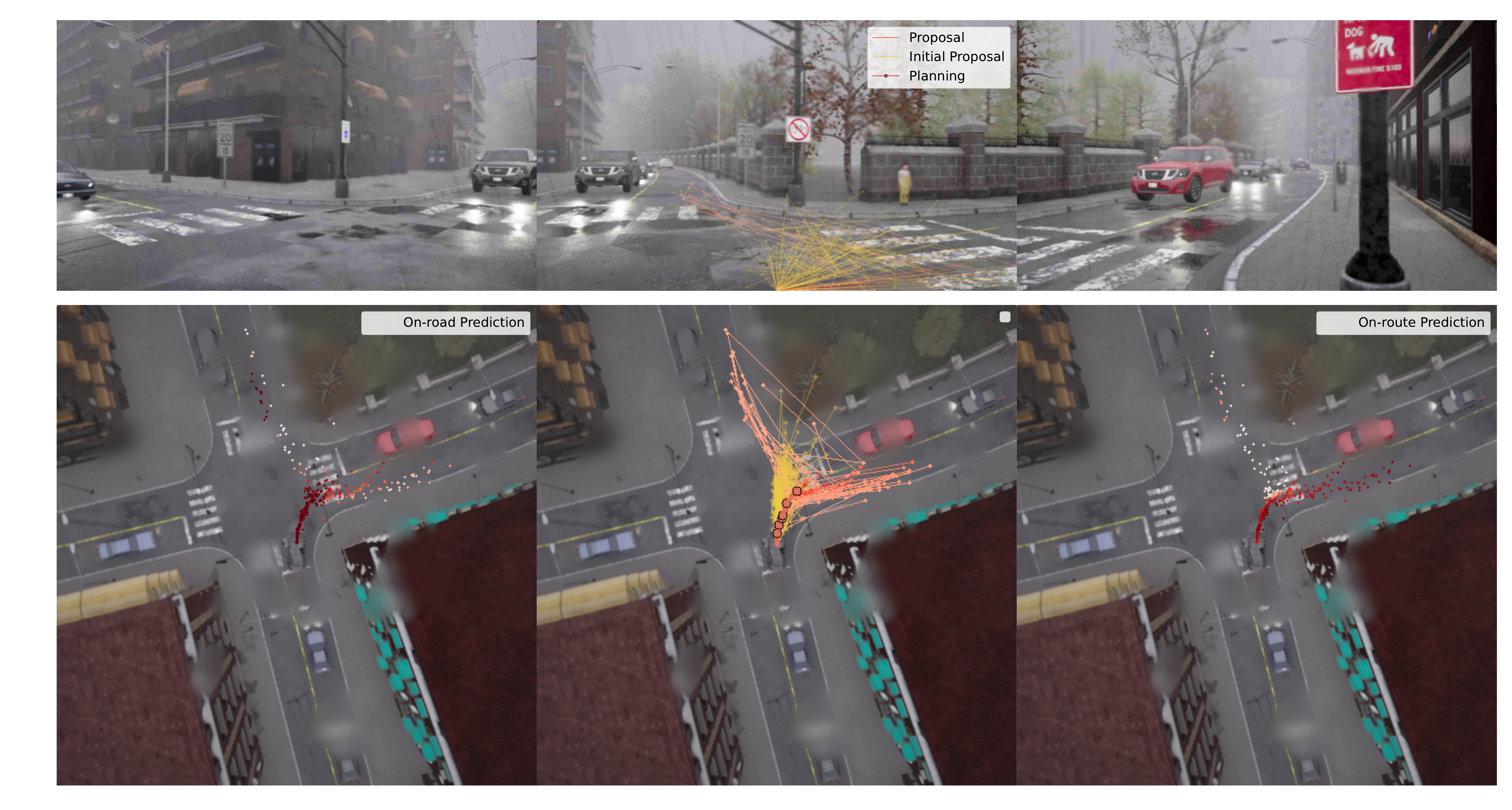}
\end{center}
\vspace{-0.3cm}
\caption{\textbf{Passability prediction results in a Bench2Drive scenario.} The lightness of the proposal lines or points decreases with their scores or predicted on-road or on-route probabilities. The proposal state is off-road if any corner point is off-road.}
\vspace{0.3cm}
\label{fig:off_road}
\end{figure*}

As shown in~\cref{fig:off_road}, \texttt{iPad} accurately predicts on-road and on-route probabilities in Bench2Drive scenarios, demonstrating awareness of both proposal heading and vehicle size. Therefore, a on-road and on-route proposal is chosen as the planning.

\newpage

\subsection{Collision Prediction}

\begin{figure*}[h]
\begin{center}
\includegraphics[width=\linewidth]{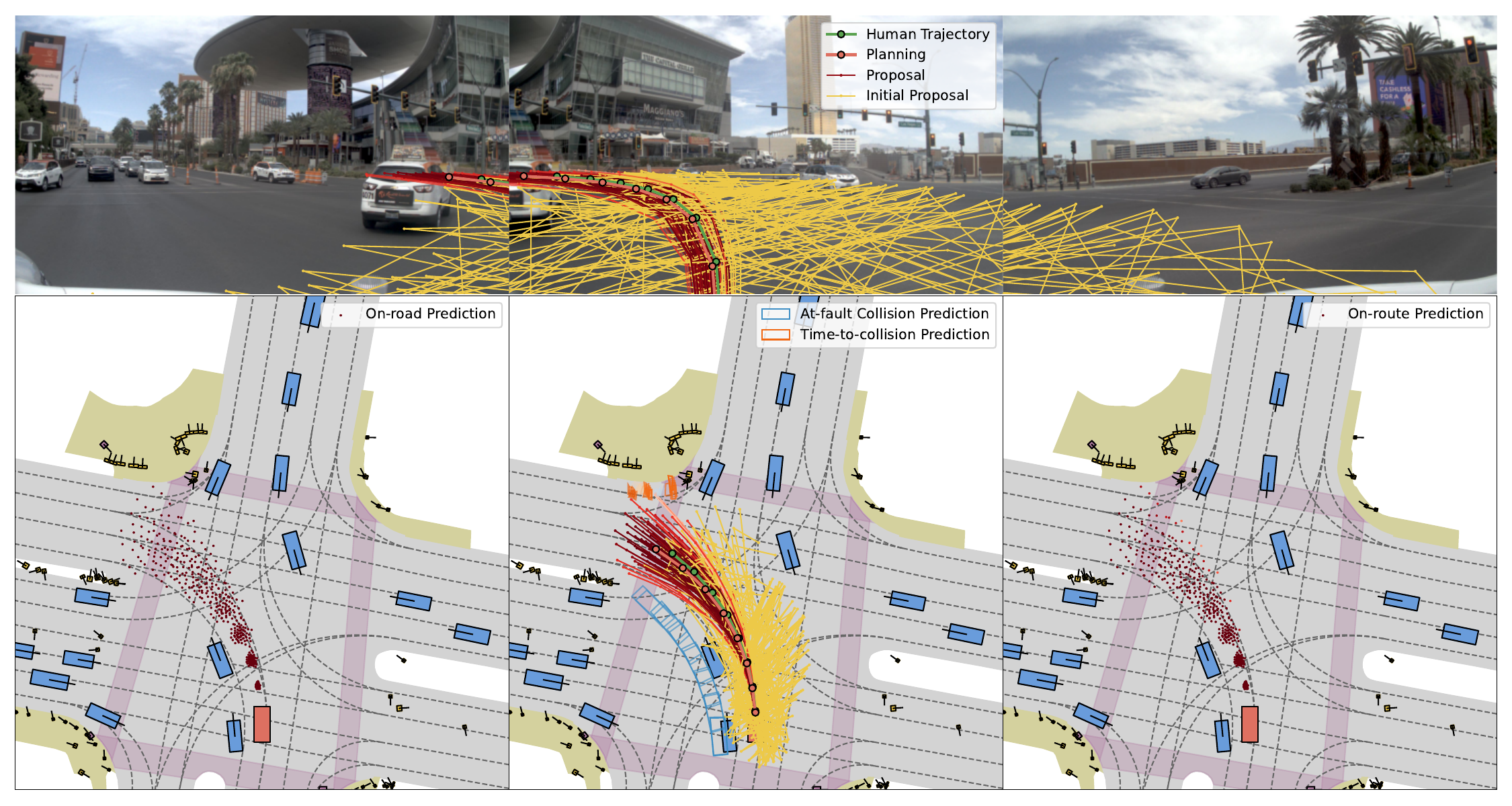}
\end{center}
\vspace{-0.3cm}
\caption{\textbf{Collision prediction results in a NAVSIM turning scenario.} The lightness of the proposal lines decreases with their scores. The lightness of the predicted agent boxes corresponds to their associated proposals.}
\vspace{0.3cm}
\label{fig:turning}
\end{figure*}

The~\cref{fig:turning} demonstrates that \texttt{iPad} can identify potential collision risks in NAVSIM scenarios by accurately predicting the future bounding boxes of at-fault and likely collided agents for outlier proposals. Consequently, the planner selects a safe centering proposal.

\begin{figure*}[h]
\begin{center}
\includegraphics[width=\linewidth]{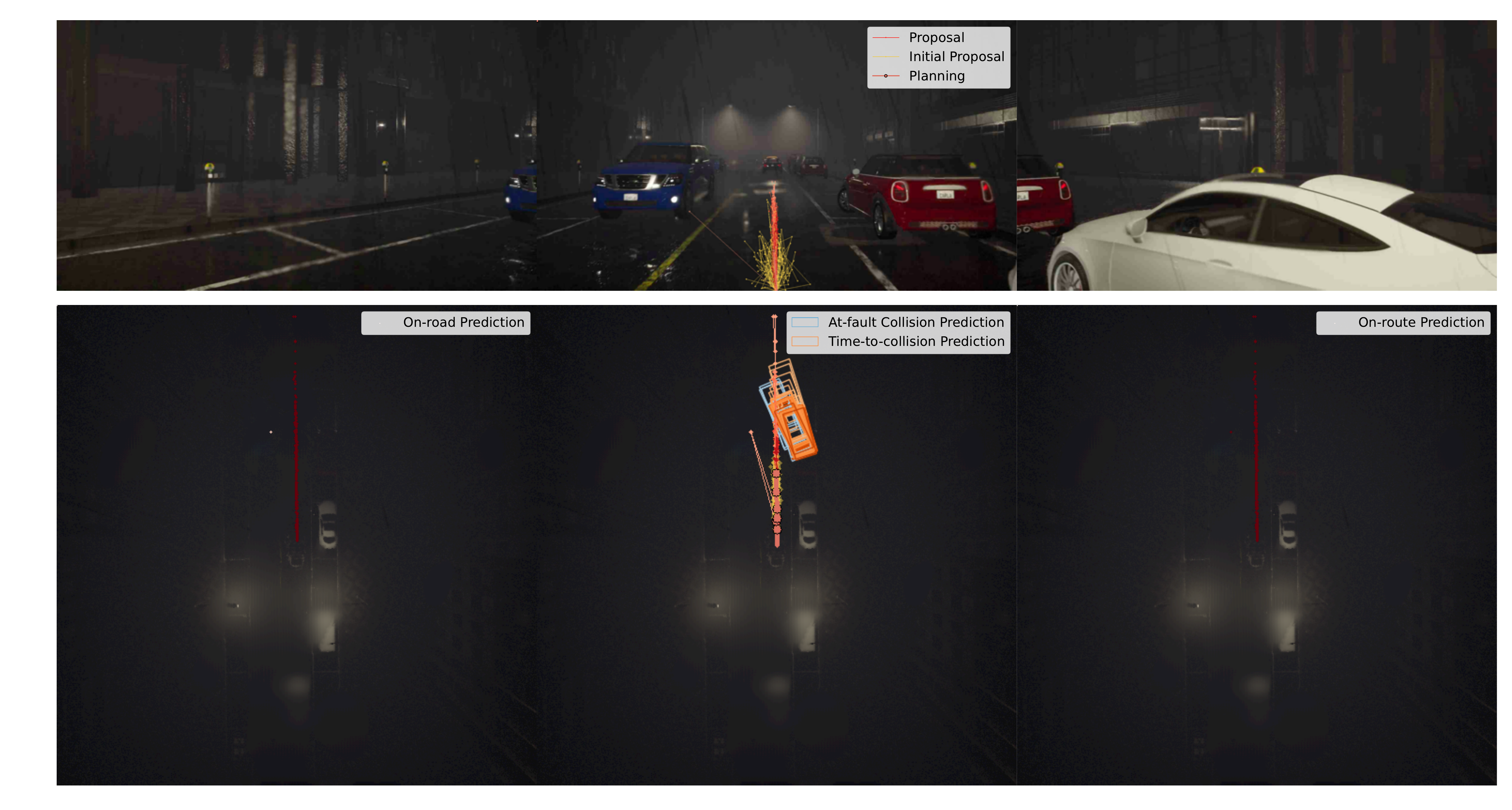}
\end{center}
\vspace{-0.3cm}
\caption{\textbf{Collision prediction results in a Bench2Drive parking cutin scenario.} The lightness of the proposal lines decreases with their scores. The lightness of the predicted agent boxes corresponds to their associated proposals.}
\vspace{-0.3cm}
\label{fig:collision}
\end{figure*}

The~\cref{fig:collision} demonstrates that \texttt{iPad} can effectively recognize potential collision risks in parking cut-in scenarios by accurately predicting the future bounding boxes of the at-fault and likely collided vehicle for dangerous proposals, when the taillights of the red car are illuminated. Therefore, a deceleration proposal is chosen as the planning.

\clearpage
\newpage
\section*{NeurIPS Paper Checklist}

\begin{enumerate}

\item {\bf Claims}
    \item[] Question: Do the main claims made in the abstract and introduction accurately reflect the paper's contributions and scope?
    \item[] Answer: \answerYes{} 
    \item[] Justification: We have clearly state the motivations and contributions for \texttt{iPad} in abstract and section~\ref{sec:intro}.
    \item[] Guidelines: The abstract and introduction in this paper accurately reflect the paper’s contributions and scope.
    \begin{itemize}
        \item The answer NA means that the abstract and introduction do not include the claims made in the paper.
        \item The abstract and/or introduction should clearly state the claims made, including the contributions made in the paper and important assumptions and limitations. A No or NA answer to this question will not be perceived well by the reviewers. 
        \item The claims made should match theoretical and experimental results, and reflect how much the results can be expected to generalize to other settings. 
        \item It is fine to include aspirational goals as motivation as long as it is clear that these goals are not attained by the paper. 
    \end{itemize}

\item {\bf Limitations}
    \item[] Question: Does the paper discuss the limitations of the work performed by the authors?
    \item[] Answer: \answerYes{} 
    \item[] Justification: We discuss the limitations of the work in \cref{sec:limitation}.
    \item[] Guidelines:
    \begin{itemize}
        \item The answer NA means that the paper has no limitation while the answer No means that the paper has limitations, but those are not discussed in the paper. 
        \item The authors are encouraged to create a separate "Limitations" section in their paper.
        \item The paper should point out any strong assumptions and how robust the results are to violations of these assumptions (e.g., independence assumptions, noiseless settings, model well-specification, asymptotic approximations only holding locally). The authors should reflect on how these assumptions might be violated in practice and what the implications would be.
        \item The authors should reflect on the scope of the claims made, e.g., if the approach was only tested on a few datasets or with a few runs. In general, empirical results often depend on implicit assumptions, which should be articulated.
        \item The authors should reflect on the factors that influence the performance of the approach. For example, a facial recognition algorithm may perform poorly when image resolution is low or images are taken in low lighting. Or a speech-to-text system might not be used reliably to provide closed captions for online lectures because it fails to handle technical jargon.
        \item The authors should discuss the computational efficiency of the proposed algorithms and how they scale with dataset size.
        \item If applicable, the authors should discuss possible limitations of their approach to address problems of privacy and fairness.
        \item While the authors might fear that complete honesty about limitations might be used by reviewers as grounds for rejection, a worse outcome might be that reviewers discover limitations that aren't acknowledged in the paper. The authors should use their best judgment and recognize that individual actions in favor of transparency play an important role in developing norms that preserve the integrity of the community. Reviewers will be specifically instructed to not penalize honesty concerning limitations.
    \end{itemize}

\item {\bf Theory assumptions and proofs}
    \item[] Question: For each theoretical result, does the paper provide the full set of assumptions and a complete (and correct) proof?
    \item[] Answer: \answerNA{}{} 
    \item[] Justification: The paper does not include theoretical results. 
    \item[] Guidelines:
    \begin{itemize}
        \item The answer NA means that the paper does not include theoretical results. 
        \item All the theorems, formulas, and proofs in the paper should be numbered and cross-referenced.
        \item All assumptions should be clearly stated or referenced in the statement of any theorems.
        \item The proofs can either appear in the main paper or the supplemental material, but if they appear in the supplemental material, the authors are encouraged to provide a short proof sketch to provide intuition. 
        \item Inversely, any informal proof provided in the core of the paper should be complemented by formal proofs provided in appendix or supplemental material.
        \item Theorems and Lemmas that the proof relies upon should be properly referenced. 
    \end{itemize}

    \item {\bf Experimental result reproducibility}
    \item[] Question: Does the paper fully disclose all the information needed to reproduce the main experimental results of the paper to the extent that it affects the main claims and/or conclusions of the paper (regardless of whether the code and data are provided or not)?
    \item[] Answer: \answerYes{} 
    \item[] Justification:  All the results in this paper can be reproduced.
    \item[] Guidelines:
    \begin{itemize}
        \item The answer NA means that the paper does not include experiments.
        \item If the paper includes experiments, a No answer to this question will not be perceived well by the reviewers: Making the paper reproducible is important, regardless of whether the code and data are provided or not.
        \item If the contribution is a dataset and/or model, the authors should describe the steps taken to make their results reproducible or verifiable. 
        \item Depending on the contribution, reproducibility can be accomplished in various ways. For example, if the contribution is a novel architecture, describing the architecture fully might suffice, or if the contribution is a specific model and empirical evaluation, it may be necessary to either make it possible for others to replicate the model with the same dataset, or provide access to the model. In general. releasing code and data is often one good way to accomplish this, but reproducibility can also be provided via detailed instructions for how to replicate the results, access to a hosted model (e.g., in the case of a large language model), releasing of a model checkpoint, or other means that are appropriate to the research performed.
        \item While NeurIPS does not require releasing code, the conference does require all submissions to provide some reasonable avenue for reproducibility, which may depend on the nature of the contribution. For example
        \begin{enumerate}
            \item If the contribution is primarily a new algorithm, the paper should make it clear how to reproduce that algorithm.
            \item If the contribution is primarily a new model architecture, the paper should describe the architecture clearly and fully.
            \item If the contribution is a new model (e.g., a large language model), then there should either be a way to access this model for reproducing the results or a way to reproduce the model (e.g., with an open-source dataset or instructions for how to construct the dataset).
            \item We recognize that reproducibility may be tricky in some cases, in which case authors are welcome to describe the particular way they provide for reproducibility. In the case of closed-source models, it may be that access to the model is limited in some way (e.g., to registered users), but it should be possible for other researchers to have some path to reproducing or verifying the results.
        \end{enumerate}
    \end{itemize}

\item {\bf Open access to data and code}
    \item[] Question: Does the paper provide open access to the data and code, with sufficient instructions to faithfully reproduce the main experimental results, as described in supplemental material?
    \item[] Answer: \answerYes{} 
    \item[] Justification: We provide open access to the data and code, with sufficient instructions to faithfully reproduce the main experimental results.
    \item[] Guidelines: 
    \begin{itemize}
        \item The answer NA means that paper does not include experiments requiring code.
        \item Please see the NeurIPS code and data submission guidelines (\url{https://nips.cc/public/guides/CodeSubmissionPolicy}) for more details.
        \item While we encourage the release of code and data, we understand that this might not be possible, so “No” is an acceptable answer. Papers cannot be rejected simply for not including code, unless this is central to the contribution (e.g., for a new open-source benchmark).
        \item The instructions should contain the exact command and environment needed to run to reproduce the results. See the NeurIPS code and data submission guidelines (\url{https://nips.cc/public/guides/CodeSubmissionPolicy}) for more details.
        \item The authors should provide instructions on data access and preparation, including how to access the raw data, preprocessed data, intermediate data, and generated data, etc.
        \item The authors should provide scripts to reproduce all experimental results for the new proposed method and baselines. If only a subset of experiments are reproducible, they should state which ones are omitted from the script and why.
        \item At submission time, to preserve anonymity, the authors should release anonymized versions (if applicable).
        \item Providing as much information as possible in supplemental material (appended to the paper) is recommended, but including URLs to data and code is permitted.
    \end{itemize}

\item {\bf Experimental setting/details}
    \item[] Question: Does the paper specify all the training and test details (e.g., data splits, hyperparameters, how they were chosen, type of optimizer, etc.) necessary to understand the results?
    \item[] Answer: \answerYes{} 
    \item[] Justification: The paper specify all the training and test details necessary to understand the results.
    \item[] Guidelines:
    \begin{itemize}
        \item The answer NA means that the paper does not include experiments.
        \item The experimental setting should be presented in the core of the paper to a level of detail that is necessary to appreciate the results and make sense of them.
        \item The full details can be provided either with the code, in appendix, or as supplemental material.
    \end{itemize}

\item {\bf Experiment statistical significance}
    \item[] Question: Does the paper report error bars suitably and correctly defined or other appropriate information about the statistical significance of the experiments?
    \item[] Answer: \answerNo{} 
    \item[] Justification: While not conducting significance tests over results, our experiments are conducted on the NAVSIM and Bench2Drive Dataset, which has a large data scale. Thus, the experimental results are stable across multiple trials, and the reported results can be
    accurately reproduced using the provided open-source code.
    \item[] Guidelines:
    \begin{itemize}
        \item The answer NA means that the paper does not include experiments.
        \item The authors should answer "Yes" if the results are accompanied by error bars, confidence intervals, or statistical significance tests, at least for the experiments that support the main claims of the paper.
        \item The factors of variability that the error bars are capturing should be clearly stated (for example, train/test split, initialization, random drawing of some parameter, or overall run with given experimental conditions).
        \item The method for calculating the error bars should be explained (closed form formula, call to a library function, bootstrap, etc.)
        \item The assumptions made should be given (e.g., Normally distributed errors).
        \item It should be clear whether the error bar is the standard deviation or the standard error of the mean.
        \item It is OK to report 1-sigma error bars, but one should state it. The authors should preferably report a 2-sigma error bar than state that they have a 96\% CI, if the hypothesis of Normality of errors is not verified.
        \item For asymmetric distributions, the authors should be careful not to show in tables or figures symmetric error bars that would yield results that are out of range (e.g. negative error rates).
        \item If error bars are reported in tables or plots, The authors should explain in the text how they were calculated and reference the corresponding figures or tables in the text.
    \end{itemize}

\item {\bf Experiments compute resources}
    \item[] Question: For each experiment, does the paper provide sufficient information on the computer resources (type of compute workers, memory, time of execution) needed to reproduce the experiments?
    \item[] Answer: \answerYes{} 
    \item[] Justification: The resources used for model training have been introduced clearly in the training section.
    \item[] Guidelines:
    \begin{itemize}
        \item The answer NA means that the paper does not include experiments.
        \item The paper should indicate the type of compute workers CPU or GPU, internal cluster, or cloud provider, including relevant memory and storage.
        \item The paper should provide the amount of compute required for each of the individual experimental runs as well as estimate the total compute. 
        \item The paper should disclose whether the full research project required more compute than the experiments reported in the paper (e.g., preliminary or failed experiments that didn't make it into the paper). 
    \end{itemize}
    
\item {\bf Code of ethics}
    \item[] Question: Does the research conducted in the paper conform, in every respect, with the NeurIPS Code of Ethics \url{https://neurips.cc/public/EthicsGuidelines}?
    \item[] Answer: \answerYes{} 
    \item[] Justification: We have all reviewed the NeurIPS Code of Ethics and striven to maintain and preserve anonymity.
    \item[] Guidelines:
    \begin{itemize}
        \item The answer NA means that the authors have not reviewed the NeurIPS Code of Ethics.
        \item If the authors answer No, they should explain the special circumstances that require a deviation from the Code of Ethics.
        \item The authors should make sure to preserve anonymity (e.g., if there is a special consideration due to laws or regulations in their jurisdiction).
    \end{itemize}

\item {\bf Broader impacts}
    \item[] Question: Does the paper discuss both potential positive societal impacts and negative societal impacts of the work performed?
    \item[] Answer: \answerYes{} 
    \item[] Justification: In the introduction, we summarize this paper’s application to autonomous driving and traffic safety.
    \item[] Guidelines:
    \begin{itemize}
        \item The answer NA means that there is no societal impact of the work performed.
        \item If the authors answer NA or No, they should explain why their work has no societal impact or why the paper does not address societal impact.
        \item Examples of negative societal impacts include potential malicious or unintended uses (e.g., disinformation, generating fake profiles, surveillance), fairness considerations (e.g., deployment of technologies that could make decisions that unfairly impact specific groups), privacy considerations, and security considerations.
        \item The conference expects that many papers will be foundational research and not tied to particular applications, let alone deployments. However, if there is a direct path to any negative applications, the authors should point it out. For example, it is legitimate to point out that an improvement in the quality of generative models could be used to generate deepfakes for disinformation. On the other hand, it is not needed to point out that a generic algorithm for optimizing neural networks could enable people to train models that generate Deepfakes faster.
        \item The authors should consider possible harms that could arise when the technology is being used as intended and functioning correctly, harms that could arise when the technology is being used as intended but gives incorrect results, and harms following from (intentional or unintentional) misuse of the technology.
        \item If there are negative societal impacts, the authors could also discuss possible mitigation strategies (e.g., gated release of models, providing defenses in addition to attacks, mechanisms for monitoring misuse, mechanisms to monitor how a system learns from feedback over time, improving the efficiency and accessibility of ML).
    \end{itemize}
    
\item {\bf Safeguards}
    \item[] Question: Does the paper describe safeguards that have been put in place for responsible release of data or models that have a high risk for misuse (e.g., pretrained language models, image generators, or scraped datasets)?
    \item[] Answer: \answerNA{} 
    \item[] Justification: The paper poses no such risks
    \item[] Guidelines:
    \begin{itemize}
        \item The answer NA means that the paper poses no such risks.
        \item Released models that have a high risk for misuse or dual-use should be released with necessary safeguards to allow for controlled use of the model, for example by requiring that users adhere to usage guidelines or restrictions to access the model or implementing safety filters. 
        \item Datasets that have been scraped from the Internet could pose safety risks. The authors should describe how they avoided releasing unsafe images.
        \item We recognize that providing effective safeguards is challenging, and many papers do not require this, but we encourage authors to take this into account and make a best faith effort.
    \end{itemize}

\item {\bf Licenses for existing assets}
    \item[] Question: Are the creators or original owners of assets (e.g., code, data, models), used in the paper, properly credited and are the license and terms of use explicitly mentioned and properly respected?
    \item[] Answer: \answerYes{} 
    \item[] Justification: The training and evaluation datasets used in this study are cited within this paper.
    \item[] Guidelines:
    \begin{itemize}
        \item The answer NA means that the paper does not use existing assets.
        \item The authors should cite the original paper that produced the code package or dataset.
        \item The authors should state which version of the asset is used and, if possible, include a URL.
        \item The name of the license (e.g., CC-BY 4.0) should be included for each asset.
        \item For scraped data from a particular source (e.g., website), the copyright and terms of service of that source should be provided.
        \item If assets are released, the license, copyright information, and terms of use in the package should be provided. For popular datasets, \url{paperswithcode.com/datasets} has curated licenses for some datasets. Their licensing guide can help determine the license of a dataset.
        \item For existing datasets that are re-packaged, both the original license and the license of the derived asset (if it has changed) should be provided.
        \item If this information is not available online, the authors are encouraged to reach out to the asset's creators.
    \end{itemize}

\item {\bf New assets}
    \item[] Question: Are new assets introduced in the paper well documented and is the documentation provided alongside the assets?
    \item[] Answer: \answerNo{} 
    \item[] Justification: The paper does not release new assets.
    \item[] Guidelines:
    \begin{itemize}
        \item The answer NA means that the paper does not release new assets.
        \item Researchers should communicate the details of the dataset/code/model as part of their submissions via structured templates. This includes details about training, license, limitations, etc. 
        \item The paper should discuss whether and how consent was obtained from people whose asset is used.
        \item At submission time, remember to anonymize your assets (if applicable). You can either create an anonymized URL or include an anonymized zip file.
    \end{itemize}

\item {\bf Crowdsourcing and research with human subjects}
    \item[] Question: For crowdsourcing experiments and research with human subjects, does the paper include the full text of instructions given to participants and screenshots, if applicable, as well as details about compensation (if any)? 
    \item[] Answer: \answerNo{} 
    \item[] Justification: The paper does not involve crowdsourcing nor research with human subjects.
    \item[] Guidelines:
    \begin{itemize}
        \item The answer NA means that the paper does not involve crowdsourcing nor research with human subjects.
        \item Including this information in the supplemental material is fine, but if the main contribution of the paper involves human subjects, then as much detail as possible should be included in the main paper. 
        \item According to the NeurIPS Code of Ethics, workers involved in data collection, curation, or other labor should be paid at least the minimum wage in the country of the data collector. 
    \end{itemize}

\item {\bf Institutional review board (IRB) approvals or equivalent for research with human subjects}
    \item[] Question: Does the paper describe potential risks incurred by study participants, whether such risks were disclosed to the subjects, and whether Institutional Review Board (IRB) approvals (or an equivalent approval/review based on the requirements of your country or institution) were obtained?
    \item[] Answer: \answerNA{} 
    \item[] Justification: The paper does not involve crowdsourcing nor research with human subjects
    \item[] Guidelines:
    \begin{itemize}
        \item The answer NA means that the paper does not involve crowdsourcing nor research with human subjects.
        \item Depending on the country in which research is conducted, IRB approval (or equivalent) may be required for any human subjects research. If you obtained IRB approval, you should clearly state this in the paper. 
        \item We recognize that the procedures for this may vary significantly between institutions and locations, and we expect authors to adhere to the NeurIPS Code of Ethics and the guidelines for their institution. 
        \item For initial submissions, do not include any information that would break anonymity (if applicable), such as the institution conducting the review.
    \end{itemize}

\item {\bf Declaration of LLM usage}
    \item[] Question: Does the paper describe the usage of LLMs if it is an important, original, or non-standard component of the core methods in this research? Note that if the LLM is used only for writing, editing, or formatting purposes and does not impact the core methodology, scientific rigorousness, or originality of the research, declaration is not required.
    \item[] Answer: \answerNA{} 
    \item[] Justification: The core method development in this research does not involve LLMs as any important, original, or non-standard components
    \item[] Guidelines:
    \begin{itemize}
        \item The answer NA means that the core method development in this research does not involve LLMs as any important, original, or non-standard components.
        \item Please refer to our LLM policy (\url{https://neurips.cc/Conferences/2025/LLM}) for what should or should not be described.
    \end{itemize}

\end{enumerate}

\end{document}